\documentclass[a4paper,10pt]{article}
\usepackage[utf8]{inputenc}
\usepackage[T1]{fontenc}
\usepackage{authblk}
\usepackage[square,numbers]{natbib}
\usepackage{color}
\usepackage{graphicx}
\usepackage{subfig}
\usepackage{amssymb}
\usepackage{latexsym}
\usepackage{url}
\usepackage{xcolor}
\usepackage{amsmath,amssymb}
\begin{document}

\title{Single Image Restoration for Participating Media Based on Prior Fusion}
\author[1]{Joel F. O. Gaya}
\author[2]{Felipe Codevilla}
\author[1]{Amanda C. Duarte}
\author[1]{Paulo L. Drews-Jr}
\author[1]{Silvia S. C. Botelho}

\affil[1]{Universidade Federal do Rio Grande, Centro de Ci\^encias Computacionais (C3), Rio Grande, Brazil}
\affil[2]{Computer Vision Center (CVC) 2, UAB building O, Barcelona, Spain}

\maketitle

\begin{abstract}
\textbf{This paper is under consideration at Pattern Recognition Letters.}
This paper describes a method to restore degraded images captured in a participating media -- fog, turbid water, sand storm, etc. Differently from the related work that only deal with a medium, we obtain generality by using an image formation model and a fusion of new image priors. The model considers the image color variation produced by the medium. The proposed restoration method is based on the fusion of these priors and supported by statistics collected on images acquired in both non-participating and participating media. The key of the method is to fuse two complementary measures --- local contrast and color data. The obtained results on underwater and foggy images demonstrate the capabilities of the proposed method. Moreover, we evaluated our method using a special dataset for which a ground-truth image is available.
\end{abstract}

%---------------------------------------------------------------------
% Introduction
%
\section{Introduction}

A \textit{participating medium} is composed of particles in suspension that affect the image formation, \textit{e.g.} underwater medium, fog, sand storm, etc. Images taken in those environments are degraded due to the interaction between light and particles. The light interacts with the medium being \textit{scattered} and \textit{absorbed}. These processes produce information loss. Furthermore, particles outside the field of view scatter over the image producing a characteristic \textit{veil}. This effect reduces the overall image contrast. 
 
Most of the restoration methods for images acquired in participating medium relies on a physical model of the image formation. The model describes a linear superposition between the signal and the veil. Usually, restoration methods estimate: (i) the properties of the veil, also named \textit{veiling light}, and (ii) the properties of the medium, \textit{i.e.} the \textit{transmission}. This is an ill-posed problem, since there are two unknown for each pixel.

Thus, many methods in the literature use multiple images \citep{Roser14,Drews15}, polarization \citep{Schechner03,Schechner05} or special hardware \citep{Narasimhan05,Fuchs08}. However, there are many practical application where just a single image is available.

\begin{figure}[!ht]
%\captionsetup[subfigure]{labelformat=empty}
\centering
\subfloat{
\includegraphics[width=38mm]{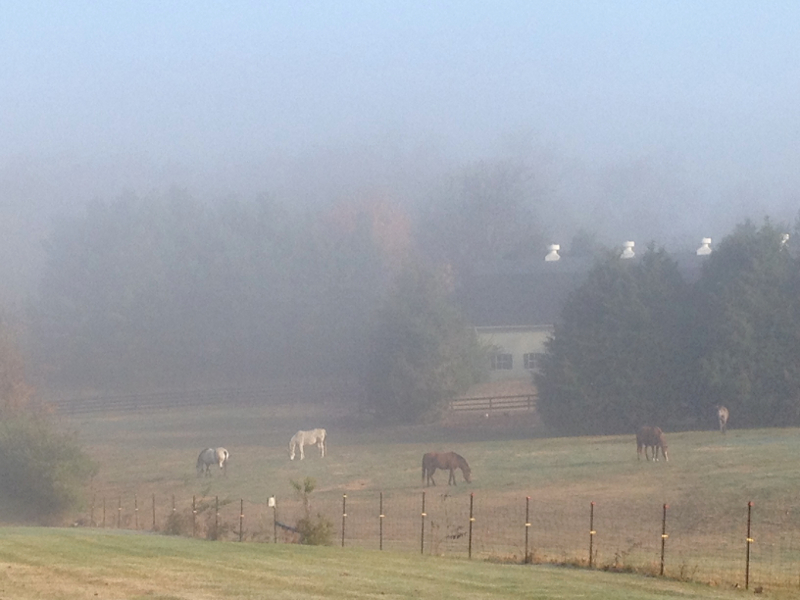}
\label{subfig:fog}}
\subfloat{
\includegraphics[width=38mm]{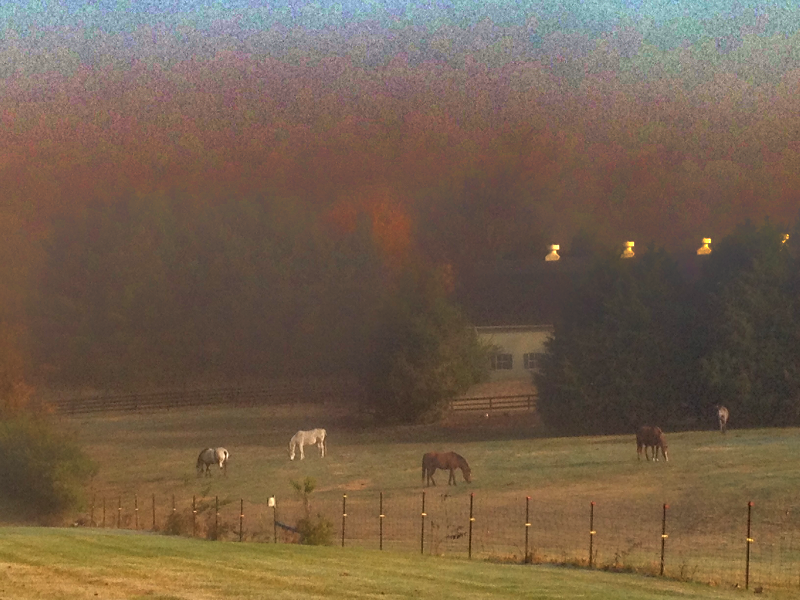}
\label{subfig:fogdcp}}
\subfloat{
\includegraphics[width=38mm]{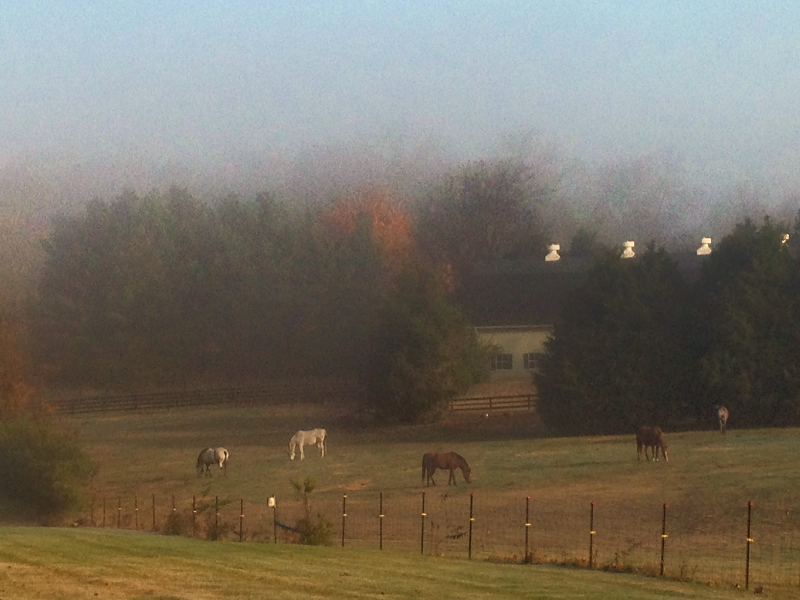}
\label{subfig:2f1}}
\caption{Failure case of a dehazing algorithms applied to a haze image (left side). The DCP \citep{He09} (center image) presents limitation when applied to 
color changes or excessive amount of noise. The proposed algorithm (right side) is able to deal with this condition.}
\label{fig:failure}
\end{figure}

The literature assumes some priors information to deal with the single image restoration. These priors helps to find a transmission and/or the veiling light allowing the solution of the ill-posed model. Most of these restoration algorithms are developed to deal with a specific participating medium. They lead to impressive results, \textit{e.g.}  the \textit{Dark Channel Prior} (DCP) \citep{He09} and the work of \cite{Fattal14} using haze images, or the work of \cite{Ancuti12} and \cite{Drews13} using underwater images. However, a simple change in the light source, or sometimes in the structure of the imaged scene, can make those method fail, \textit{e.g.} Fig. \ref{fig:failure}.

We believe that a more general estimation of the characteristics of the medium is required to obtain a robust method to deal with environmental changes.

In this context, we propose an automatic single image restoration method designed to deal with images acquired in \textit{participating media}, mainly underwater or hazy condition. We obtain generality by using (i) physical model, (ii) a robust estimation of the medium parameters, and (iii) the fusion of  complementary \textit{novel} priors. These characteristics enable us to successfully restore both hazy and underwater images.

Finally, we list the following contributions for this paper:

\begin{itemize}
    \item A new general color prior to be adopted in participating medium. The prior is named \textit{Veil Difference Prior}. Interestingly, this prior can be seen as a generalization of the DCP \citep{He09} for participating medium (Sec. \ref{sec:rest}). Furthermore, we proposed a novel contrast prior based on the physical model called \textit{Contrast Prior}.
    
     \item A fusion strategy to estimate the transmission. We show this approach allows us to obtain a better transmission estimation.
     
    \item Results are validated in quantitative terms. Differently from the state-of-the-art that uses qualitative or quantitative evaluation based on empirical metrics, we use a dataset acquired in a controlled underwater environment where the amount of degradation is previously known.
\end{itemize}

% Close Introduction
%---------------------------------------------------------------------

%---------------------------------------------------------------------
% Image Formation Model
%

\section{Image Formation Model}
\label{sec:model}

The light propagating through a participating medium suffer \textit{scattering} and \textit{absorption} by the suspended particles. The image \textit{signal}, \textit{i.e.} the imaged scene, suffers from \textit{attenuation} where just a small portion of the light reaches the camera. \textit{Forward scattering} occurs when the light rays coming from the scene are scattered in small angles. It creates a blurry effect on the image, mainly underwater. However, this effect presents a small contribution to the total image degradation and is frequently neglected \cite{Schechner05}. The \textit{backscattering} results from the interaction between the illumination sources and particles dispersed in the medium. It creates a characteristic veil on the image which reduces the contrast and attenuates the signal information. We define an image captured in a participating medium as:
\begin{equation}
I_\lambda(x) = E_{d_\lambda}(x) + E_{bs_\lambda}(x), \label{eq:initial}
\end{equation}
\noindent where each color channel $\lambda \in \{R,G,B\}$, $E_{d_\lambda}(x) $ is the
direct component (signal) and  $E_{bs_\lambda}(x)$
is the backscattering component. The rest of this section explains each components of
Eq. (\ref{eq:initial}).

%---------------------------------------------------------------------------
\subsection{Direct Component}

The direct component, $E_{d_\lambda}(x)$, is defined as: 

\begin{equation}
E_{d_\lambda}(x) = J_{\lambda}(x) \thinspace e^{-cd(x)}, \label{eq:direct}
\end{equation}

\noindent where  $J_{\lambda}(x)$ is the signal with no degradation, which is attenuated by $e^{-cd(x)}$, named as transmission $t(x)$. We assume that, considering $J_\lambda(x)$ as a general image taken from a Lambertian surface, $J_\lambda(x)$ can also be described by the color constancy image formation model \citep{Van07}:
\begin{equation}
J_\lambda(x) = L_\lambda(x) \thinspace M_\lambda(x) \thinspace C_\lambda(x), \label{eq:ed}
\end{equation}
\noindent where $L_\lambda(x)$ is the light source, $M_\lambda(x)$
is the reflectivity of the imaged object, and 
$C_\lambda(x)$ are the camera parameters. Omitting the camera
parameters and considering the light source as constant, we have that:
\begin{equation}
J_{\lambda}(x)  =  L_{\lambda} \thinspace M_{\lambda}(x). 
\label{eq:light}
\end{equation}

In images acquired in participating medium, we consider that natural light comes from a limited cone above the scene, the \textit{optical manhole cone} \citep{cronin01}. 
For this reason, we assume that the light source is related to the cone size and
also influenced by the environment.
%-----------------------------------------------------------------------

\subsection{Backscattering Component}

Following the the model proposed by Jaffe-McGlamery \citep{Jaffe90,McGlamery80} and the simplification proposed by \cite{Schechner05}, the backscattering component, $E_{bs_\lambda}(x)$, can be defined as:
\begin{equation}
E_{bs_\lambda}(x) = A_\lambda^{D} \cdot \thinspace (1 - t(x)), \label{eq:bs1}
\end{equation}
\noindent where $A_\lambda^{D}$ is the veiling light, here also called ambient light constant, that represents the color and radiance characteristics of the media. This constant is related to the \textit{optical manhole cone} placed above the $LOS$ (Line of Sight). Also, the veiling light is associated with the distance between the light source and scene ($d$) and influenced by the environment. Term $(1 - t(x))$ weights the effect of the backscattering as a function of the distance, $d(x)$, from the object to the camera. The longer the distance, the higher the chance that $A_\lambda^{D}$ scatters over the scene.
%-----------------------------------------------------------------------

\subsection{Final Model}

Assuming a large distance between the light source and a small variation in the scene depth ($D$), we can consider that the $LOS$ and the objects are \textit{all} illuminated by the same light source. The main assumption of this section is the ambient light, $A_{\lambda}^{D}$, and the constant light source, $L_\lambda$, can be approximated by the same constant. Thus, Eqs. (\ref{eq:direct}) and (\ref{eq:light}) allow us to define the final model as:
\begin{equation}
I_\lambda(x) = A_\lambda^{D}  \cdot \thinspace M_\lambda(x)  \cdot \thinspace t(x) + A_\lambda^{D}  \cdot \thinspace (1 - t(x)). \label{eq:final}
\end{equation}

Note that this equation presents a generalization from the \textit{Koschmieder}'s equation \citep{Kosch24} that is commonly adopted by dehazing methods, \textit{e.g.} \citep{He09} and \citep{Fattal08}. Assuming a intensity values in the interval $[0;1]$, Eq. (\ref{eq:final}) turns into the \textit{Koschmieder}'s equation when $A_\lambda^D \approx [1,1,1]$ (approximately
white) . In Section \ref{sec:evaluation}, we show that can be jointly obtain color recovering and haze removal by using Eq. (\ref{eq:final}). 

% Close Image Formation Model
%---------------------------------------------------------------------

%---------------------------------------------------------------------
% Related Works
%
\section{Related Works}
\label{sec:related}

Both transmission and ambient light must be estimated to obtain the true object color reflectivity. For this, we need to assume some properties that a haze-free image should have, \textit{i.e.}, an image without ambient interference. These properties are usually image priors, or assumptions that can be used to find indicators of the amount of turbidity of a certain image.

\citet{Fattal08} assumed that there is no covariance between the reflectance and the illumination. Thus, the transmission can be defined as the source of covariance. However, it has been shown that the assumption works only for low degradation conditions \citep{Fattal14}.

\cite{Fattal14} also proposed a method to estimate the transmission
that uses a \textit{color line assumption} \citep{Omer04} . It infers the transmission by finding the intersection point between
the color line and the vector with the orientation of the veiling light. The success of the method depends on finding patches were some model properties exist.

One of the most important method is based on the Dark Channel Prior \citep{He09}, where the minimum value of the image channels in a small image patch provides an indication of the transmission, $t(x)$. The method obtains good results, but it works only for white colored haze and does not work well for underwater environments \citep{Drews13}. 

The same method has been adapted several times for underwater environments, \textit{e.g.} \cite{Chiang12}, \cite{Codevilla13}, \cite{Galdran15}, \cite{Lu15}, and \cite{Drews16}. However, all adaptations lacks to consider the large range of colors that exist underwater by assuming some specific conditions such as the blue channel predominance \citep{Chiang12} or the red channel absorption \citep{Galdran15,Drews16}. Our method (Section \ref{sec:model}) do not consider the participating media as having single color properties. We show this assumption is helpful in achieving accurate image restoration from any type of environment, mainly underwater.

There are also approaches that directly manipulate some image properties, \textit{e.g.} contrast, blur, and noise, in order to try to improve them. Many general image enhancing methods can be used to recover the visibility through turbid media, \textit{e.g.} CLAHE \citep{Hummel77}, Bilateral Filter \citep{Tomasi98}, and Color Constancy \citep{Van07}. There are examples of enhancement method fusions in the literature, \textit{e.g.} \citep{Ancuti12} and \citep{Bazeille2006}.The direct manipulation of the image properties reduces the haze at a cost of also degrading some of the desired image properties. Moreover, enhancement methods do not usually consider the depth variation that exists in images acquired in participating media.
% Close Related Works
%---------------------------------------------------------------------

%---------------------------------------------------------------------
% Composite Transmission Estimation
%
\section{Composite Transmission Estimation}
\label{sec:composite}

The use of a single indicator, such as the color \citep{He09,Chiang12}, or the contrast \citep{Tan08}, is decisive but not sufficient for most situations. For instance, it is not possible to know if a signal corresponds to an object reflectance of a given color, or if it presents a certain color due to the ambient light. The same problem occurs in the image contrast, \textit{i.e.} it is not possible to know if a image area presents weak contrast or if this contrast is already attenuated by the medium. Thus, we propose the \textit{Composite Transmission} by fusing two novel image priors, the \textit{Veil Difference Prior} and the \textit{Contrast Prior}.

%---------------------------------------------------------------------
\subsection{Veil Difference Prior}

\begin{figure}[!ht]
\centering
\subfloat{
\includegraphics[trim={0cm 0cm 1cm 1cm},width=55mm]{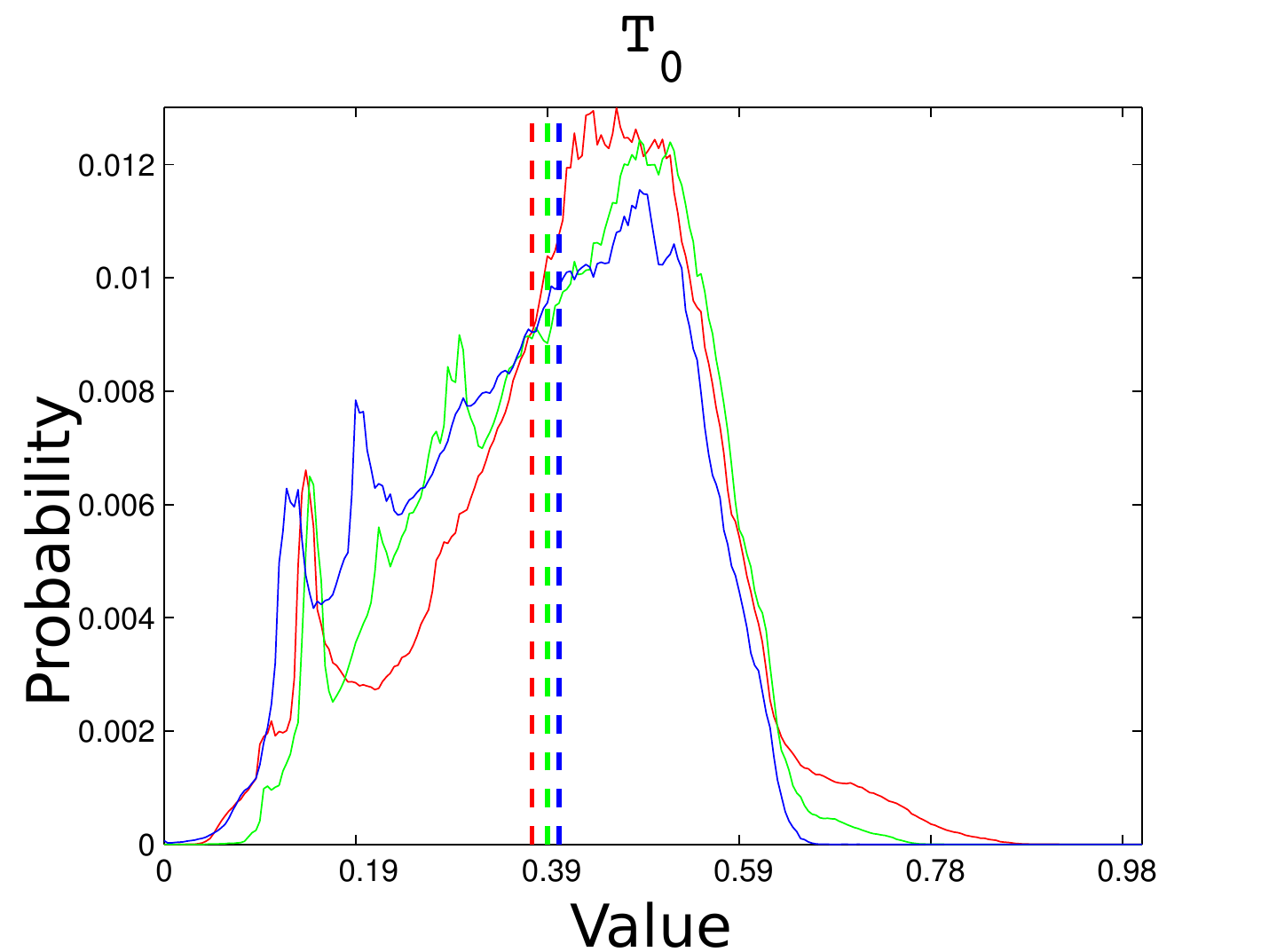}
\label{subfig:1f1}
}
\subfloat{
\includegraphics[trim={0cm 0cm 1cm 1cm},width=55mm]{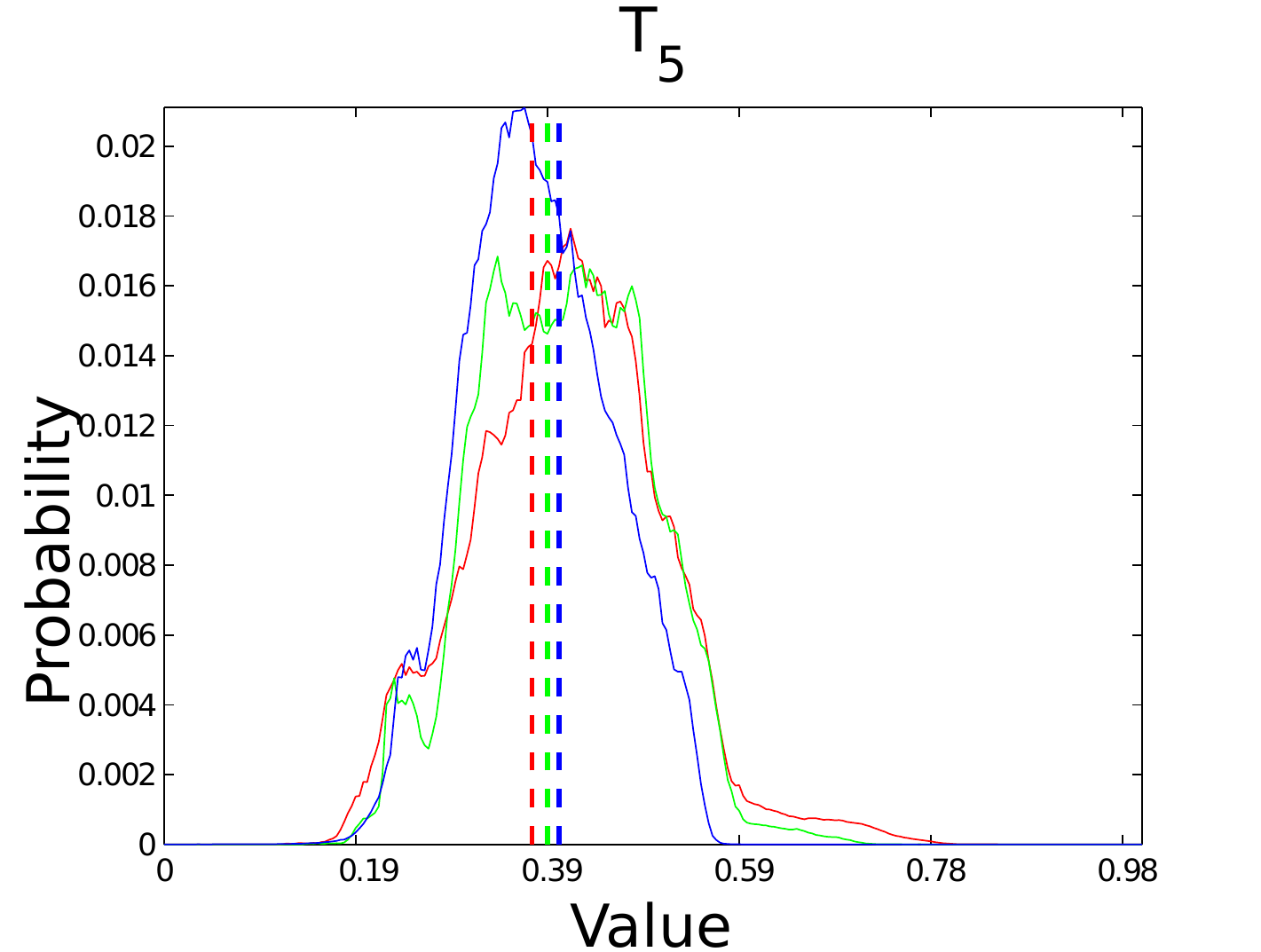}
\label{subfig:1f2}
}

\subfloat{
\includegraphics[trim={0cm 0cm 1cm 1cm},width=55mm]{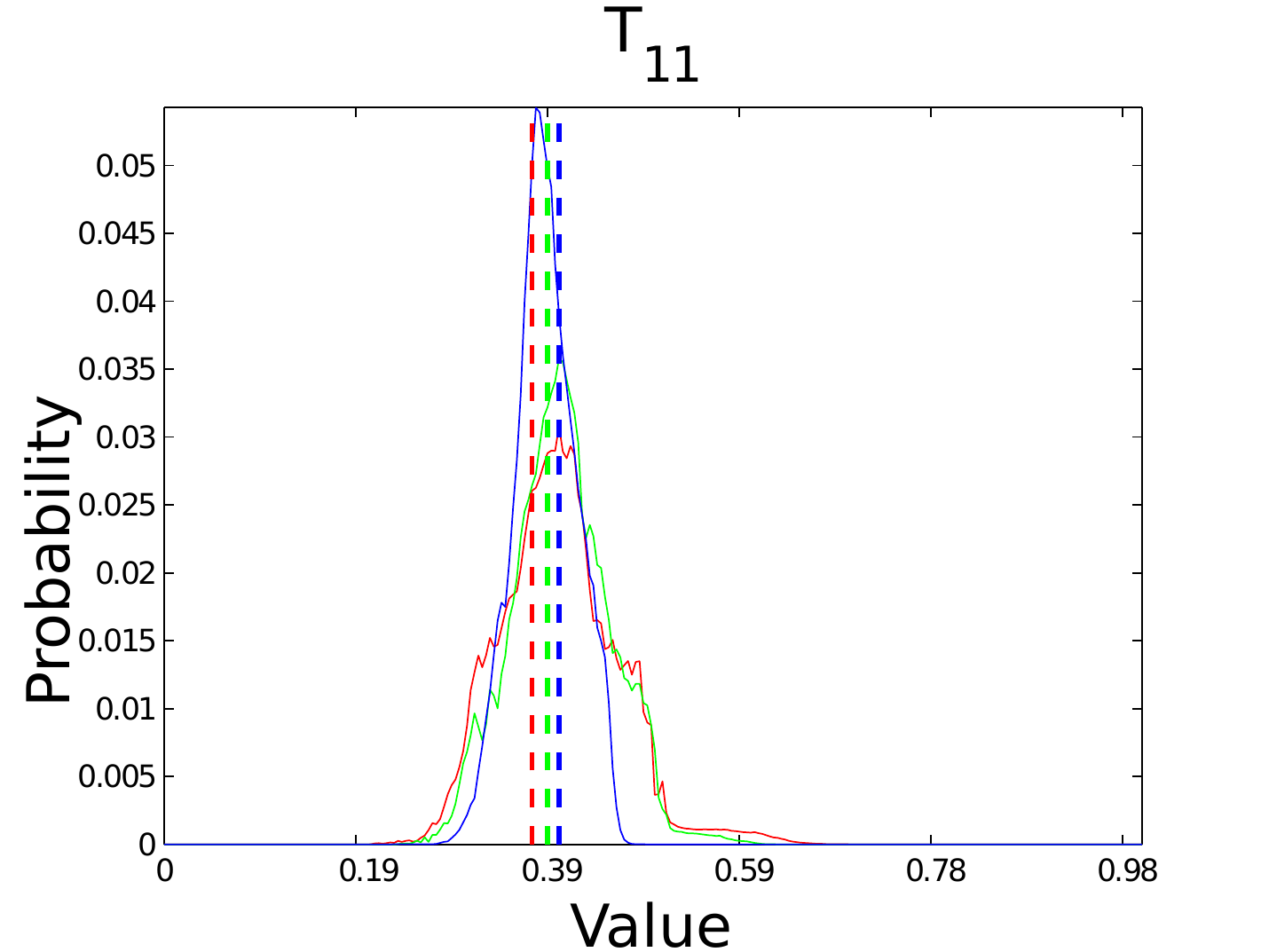}
\label{subfig:2f3}
}
\subfloat{
\includegraphics[trim={0cm 0cm 1cm 1cm},width=55mm]{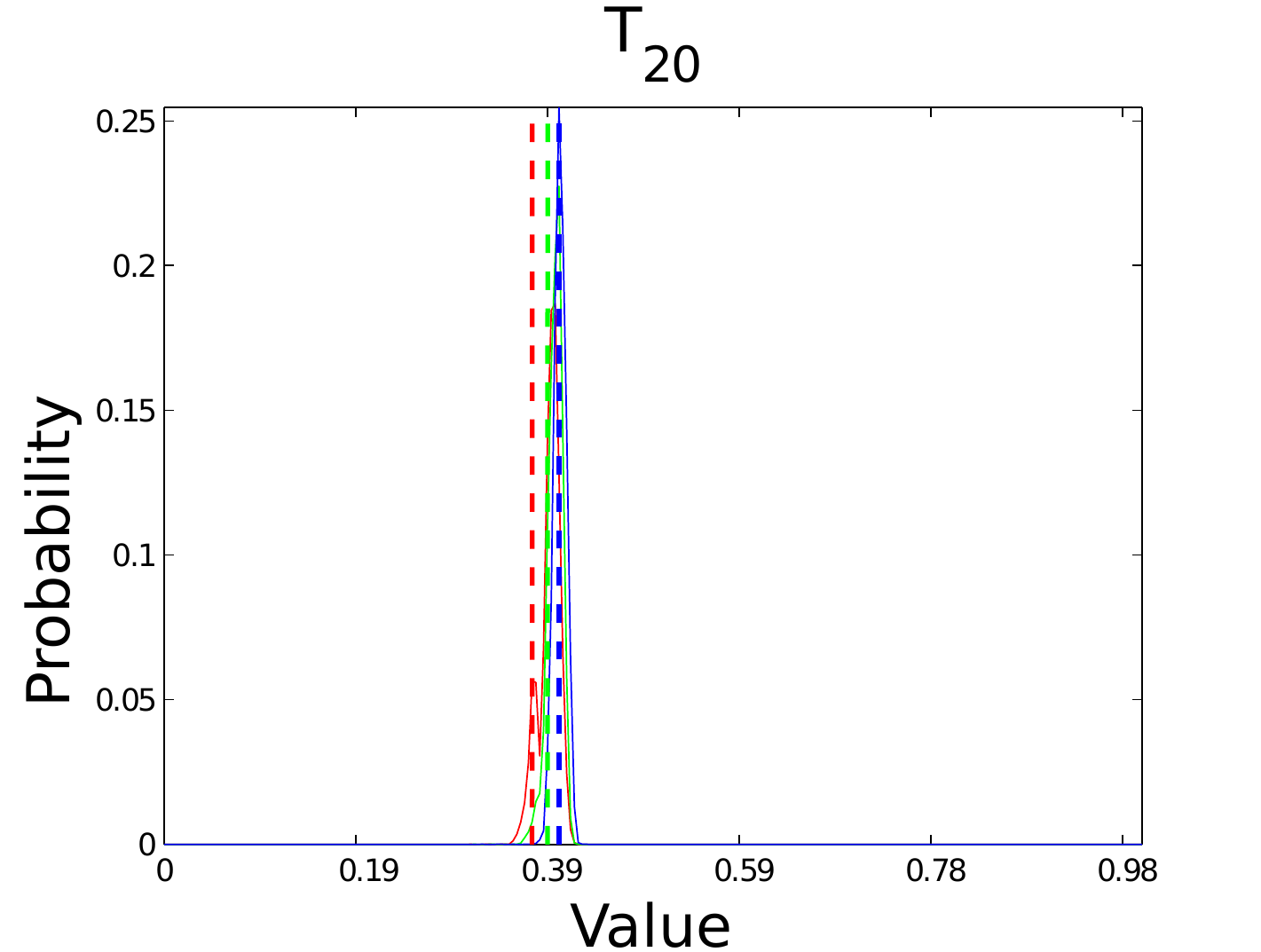}
\label{subfig:0f4_}
}
\caption{Histograms of a controlled underwater scene captured under several
levels of degradation. $T_0$ is an image acquired in a clean water. $T_5$, $T_{11}$ and $T_{20}$ present three levels of degradation created through
the addition of milk on the water \citep{duarte2016dataset}. The ambient light is represented by the dashed lines. The color of the lines represents the three RGB channels of the images. Fig. \ref{fig:turbid} shows the images adopted to obtain the histograms.}
\label{fig:hist}
\end{figure}

Based on the image formation model observation, we stated the following assumption: the image pixel color tends to be closer to the ambient light constant when the image is degraded by the participating medium. Fig. \ref{fig:hist} shows the histogram of a controlled underwater scene imaged under several levels of degradation. It can be seen that the intensities of the pixels tend to be closer to the ambient light constant (represented by dashed lines) as the level of degradation increases. Eq. \ref{eq:final} shows that the image pixel color tends to the ambient light constant with the increasing of the optical depth $cd$.

We assume all pixels from a small patch $\Omega(x)$ centered at $x$ are equally distanced to the camera. This assumption is valid since the distance to the camera is usually larger than the distance between objects in the scene. Since it is not always valid, we estimate only a rough transmission using the veil difference as:
\small{
\begin{equation}
\tilde{t}_v(x)  = \frac{\underset{ \kern -2.5pc \lambda \in \{r,g,b\} \ \  y  \in \Omega(x)}{\max(  \ \max(|I_\lambda(y)-A_{\lambda}^{D}|))}}{\underset{ \kern -2.5pc \lambda \in \{r,g,b\} \ \  y \in \Omega(x)}{\max( \ \max(|J_\lambda(y)-A_{\lambda}^{D}|))} }.     \label{eq:veilinitial}
\end{equation}
}
\normalsize
We take the channel with the highest difference since it is the one with more information.
The problem is the haze-free image is usually unknown. However, we perceive that they tend to have a significant difference to the light source on at least one of its color channels when looking at haze-free images. Formally, for an image $J_\lambda$, we define the Veil Difference Prior as:
\begin{equation}
\underset{ \kern -3pc \lambda \in \{r,g,b\} \  y \in \Omega(x)}{\max( \ \max(|J_\lambda(y)-A_{\lambda}^{D}|)) } =  \underset{ \kern -4pc \lambda \in \{r,g,b\}}{\max( \ \max(1-A_{\lambda}^{D}, A_{\lambda}^{D}))  }.
\label{eq:veilprior}
\end{equation}

\normalsize
We compute the average histogram and other priors on 2,000 haze-free images (Fig. \ref{subfig:veil} ) and other 2,000 images acquired in participating media (Fig. \ref{subfig:veilpm}) to evaluate this prior. The obtained result shows that the prior is clearly not as strong as presented by \cite{He09} for haze-free images. However, the average for the prior on degraded participating media showed a very high range of values indicating a sensitivity to the presence of the medium.

Finally, the Veil Difference Prior (VDP) can be understood as a \textit{generalization} of the dark channel prior from \cite{He09}. The Eq. (\ref{eq:veilprior}) turns into  $\max(\max(1-{J_\lambda(x)})) = 1$ for $\lambda \in \{r,g,b\}$ which is equivalent to the dark channel prior by assuming the ambient light color as pure white, \textit{i.e.} $A_\lambda^D = [1, 1, 1]$. 

Using the VDP we can compute the transmission by replacing Eq. (\ref{eq:veilprior}) into Eq. (\ref{eq:veilinitial}):

\begin{equation}
\tilde{t}_v(x)  = \frac{ \underset{ \kern -2.5pc \lambda \in \{r,g,b\} \ \  y \in \Omega(x)} {\max(  \ \max(|I_\lambda(y)-A_{\lambda}^{D}|))}}{\underset{ \kern -5.5pc \lambda \in \{r,g,b\}} {\max( \ \max(1-A_{\lambda}^{D}, A_{\lambda}^{D}))} }.
\label{eq:chromaFinal}
\end{equation}
\normalsize
 
We show the Veil Difference Transmission on Fig. \ref{subfig:tv}. One can notice that this transmission captures the depth variation on the image. 
%---------------------------------------------------------------------

\subsection{Contrast Prior}

We can observe on Fig. \ref{fig:hist} that, besides the color approximation to veiling light, we have shrinking into the histogram shape, dramatically reducing the global image contrast. Thus, we assume  the following local indicator: for a given image patch $I_\lambda$, the contrast on this patch tends to reduce proportionally to an increase at the presence of the medium. Thus, the contrast transmission can be computed as:

\begin{equation}
\tilde{t}_c(x)  = \frac{\underset{ \kern -5pc \lambda \in \{r,g,b\} \ \  y \in \Omega(x)}{\max(  \ \max(I_\lambda(y)) - \min(I_\lambda(y)))}}{\underset{ \kern -5pc \lambda \in \{r,g,b\} \ \  y \in \Omega(x)}{\max( \ \max(J_\lambda(y)) - \min(J_\lambda(y)))} },    \label{eq:contrast}
\end{equation}

\normalsize 
\noindent this ratio, analogously to Eq. (\ref{eq:veilinitial}), represents the lost of contrast information due to the medium. However, the contrast of the haze-free version of image is not available. So, we assume that haze-free image patches present the maximum possible range defining the Contrast Prior as:

\begin{equation}
\underset{ \kern -6pc \lambda \in \{r,g,b\} \ y \in \Omega(x)}{\max( \ \max(J_\lambda(y)) - \min(J_\lambda(y)))} = 1.       \label{eq:rangeprior}
\end{equation}

For the contrast Prior we also made an statistical evaluation on a large dataset (Fig.\ref{fig:histo}).
Figure \ref{subfig:con} shows a significant number of patches have value one. Also, it presents a similar behavior as the VDP and UDCP (Fig. \ref{subfig:conpm}).
Thus, we have the contrast transmission computed as:

\begin{equation}
\tilde{t}_c(x)  = \underset{ \kern -6pc \lambda \in \{r,g,b\} \ \  y \in \Omega(x)}{\max(  \ \max(I_\lambda(y)) - \min(I_\lambda(y)))}. \label{eq:contrasttrans}
\end{equation}

We show the contrast transmission on Fig. \ref{subfig:tc}. One can perceive that this transmission is sparse but presents also the depth variation.
%---------------------------------------------------------------------

\subsection{Refining Transmission}

When the transmission is computed over a patch, $\Omega(x)$, it produces a coarse estimation. Then, the transmission map for $\tilde{t}_v(x)$ and $\tilde{t}_c(x)$,
 must be refined. We choose to use the soft matting algorithm \citep{Levin08}.

Many other refinement algorithms have been applied with smaller computational costs \citep{Fattal14} or \citep{Galdran15}. However, \cite{Levin08} not solely maximizes the agreement of the transmission with edges and corners of the image. Also, it tends to spread the transmission when there is discontinuities in the rough transmission $\tilde{t}$. This maximization is specially effective when there is more sparse measures such as the Contrast Prior (Fig. \ref{subfig:tc} and Fig. \ref{subfig:tcref}).

%---------------------------------------------------------------------
% Composite Transmission
%
\subsection{Composite Transmission}

We define that the transmission can be computed by joining
multiple priors by a function:

\begin{equation}
    t(x) = f(t_1(x), t_2(x), ..., t_n(x)). \label{eq:prior fusion}
\end{equation}

\begin{figure}[!ht]
\centering
\subfloat[DCP \textit{H.F.}]{
\includegraphics[ trim={0cm 0.5cm 0cm 0.5cm},width=30mm]{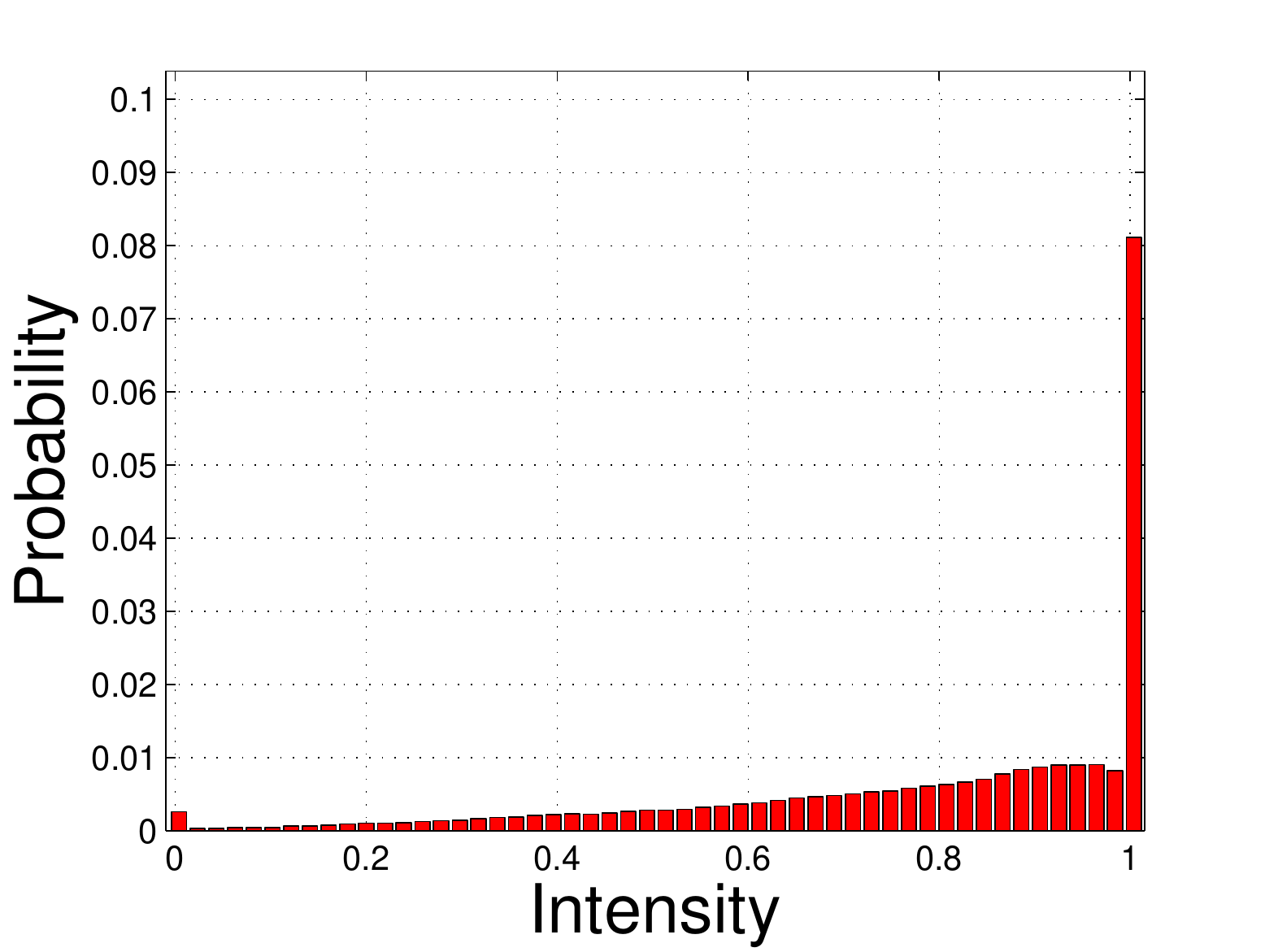}
\label{subfig:dcp}
}
\subfloat[DCP \textit{P.M.}]{
\includegraphics[ trim={0cm 0.5cm 0cm 0.5cm},width=30mm]{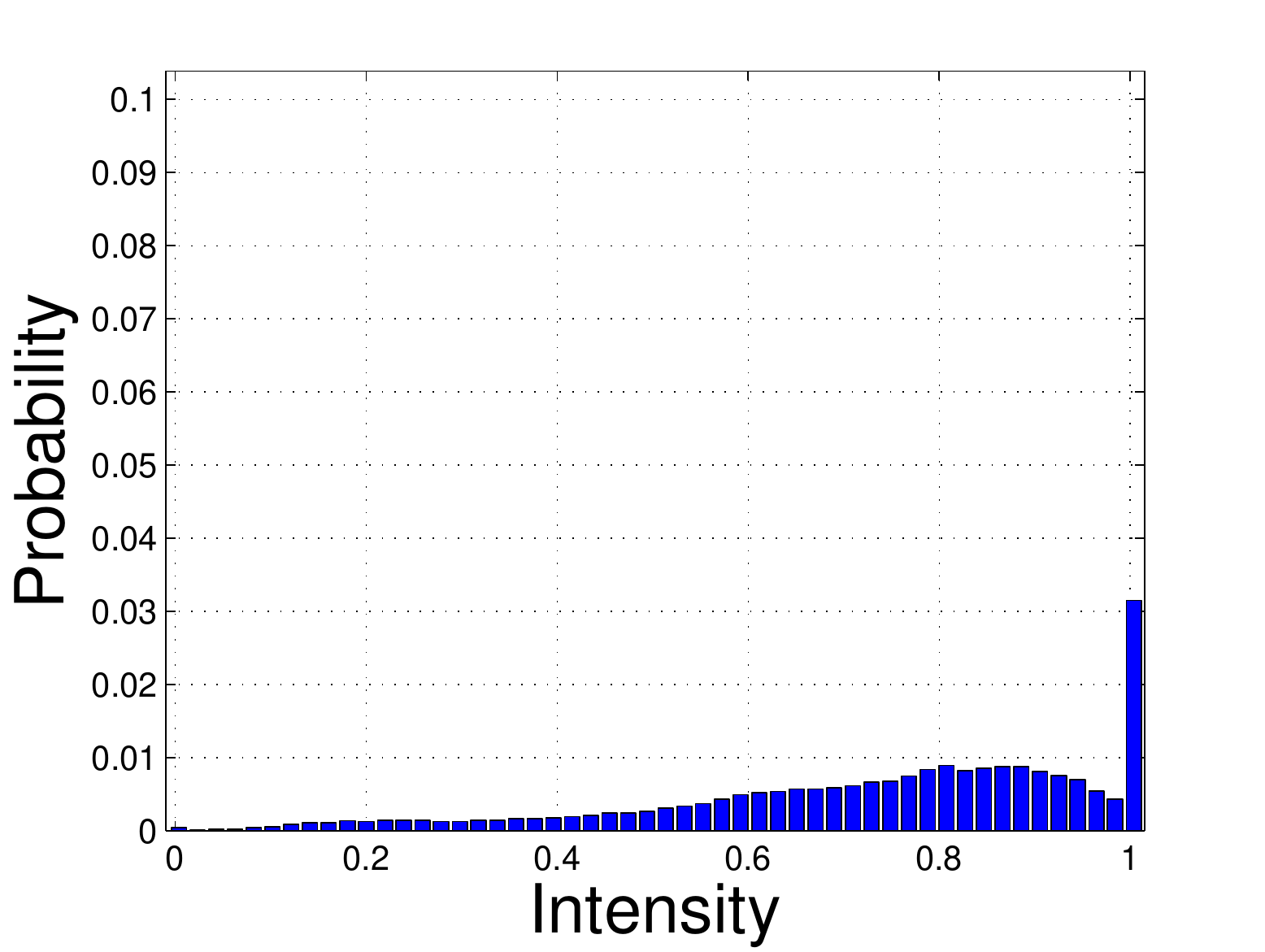}
\label{subfig:dcppm}
}\subfloat[UDCP \textit{H.F.}]{
\includegraphics[ trim={0cm 0.5cm 0cm 0.5cm},width=30mm]{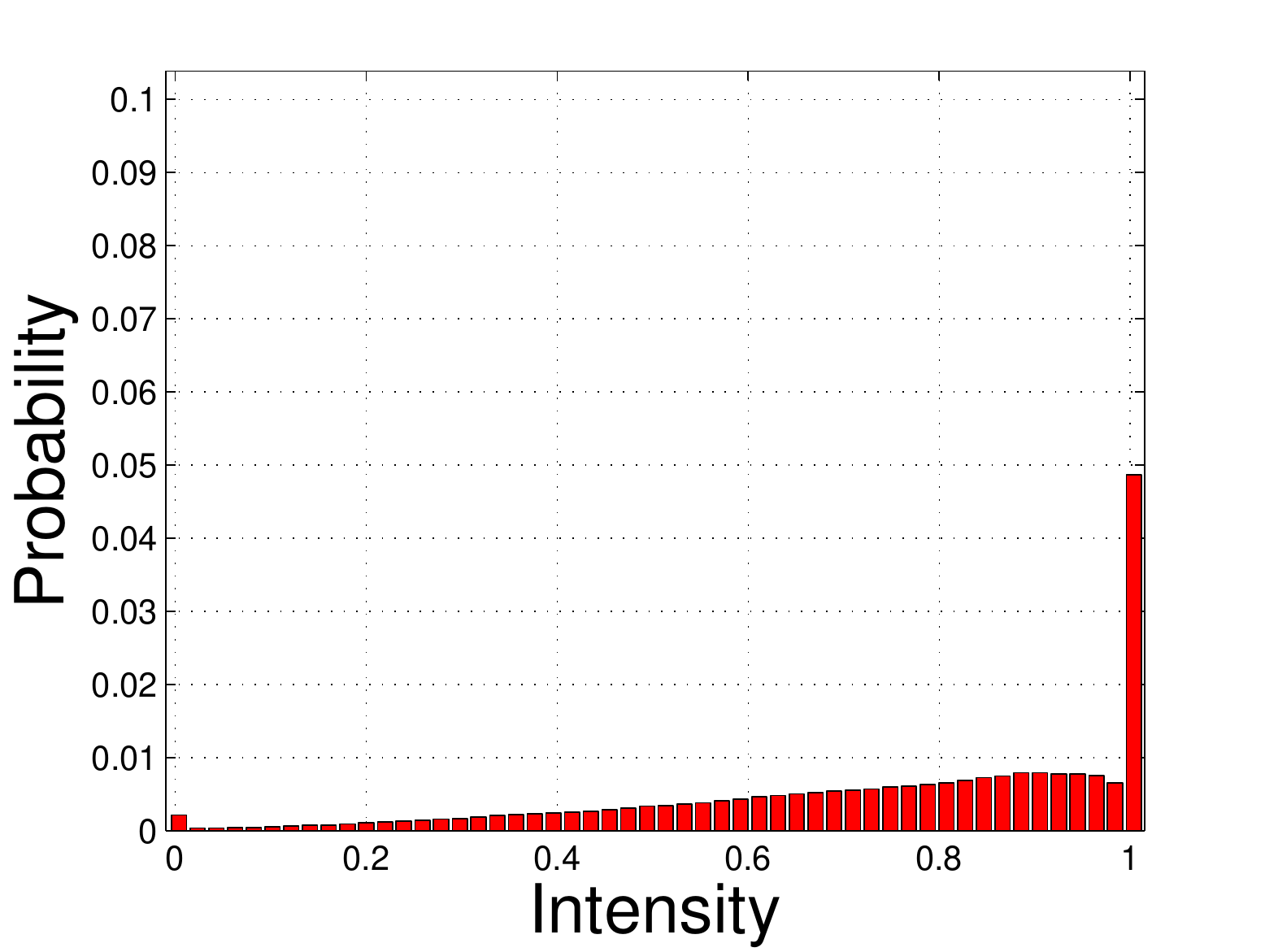}
\label{subfig:dcpgb}
}
\subfloat[UDCP \textit{P.M.}]{
\includegraphics[ trim={0cm 0.5cm 0cm 0.5cm},width=30mm]{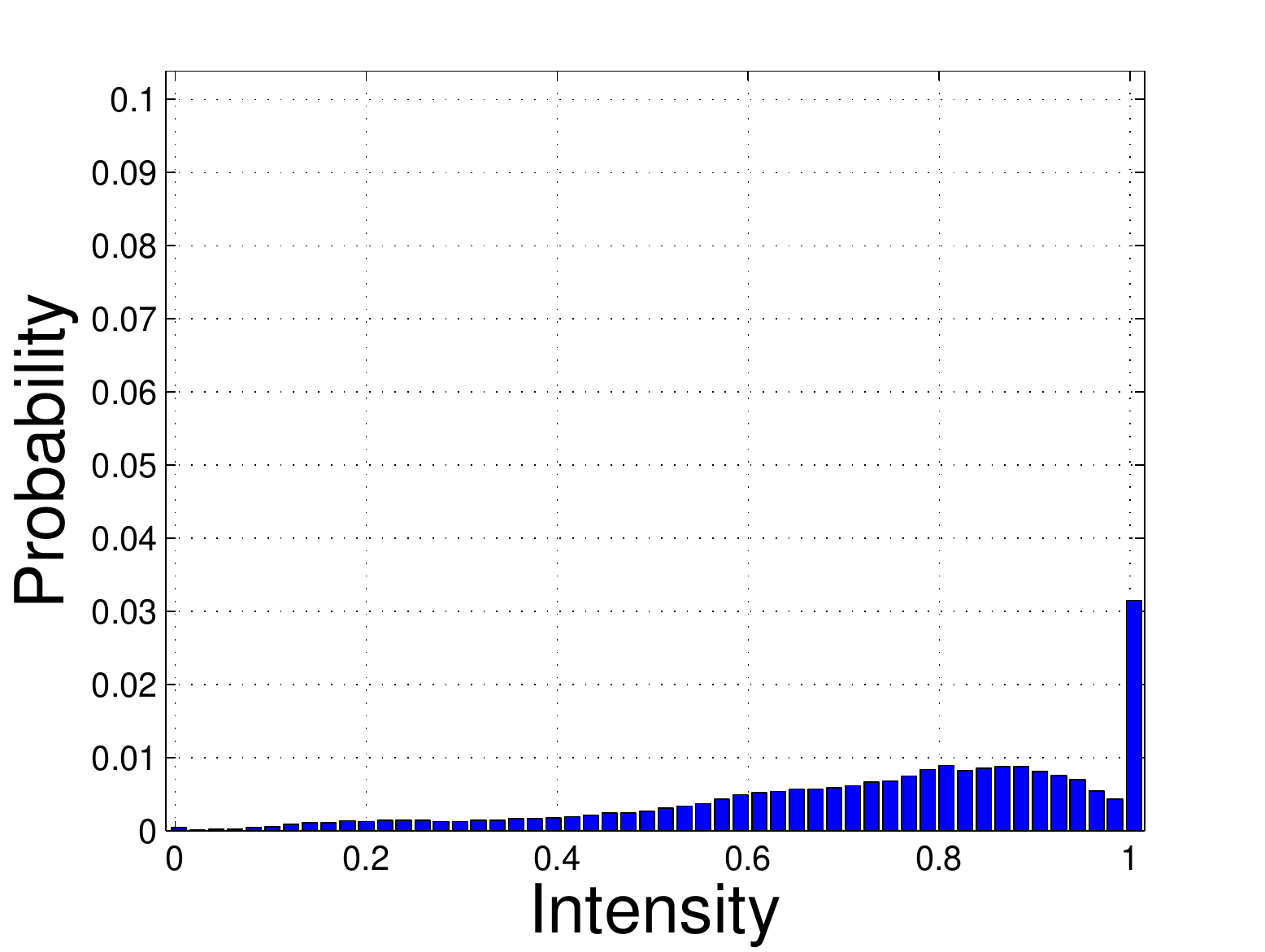}
\label{subfig:dcpgbpm}
}

\subfloat[VDP \textit{H.F.}]{
\includegraphics[ trim={0cm 0.5cm 0cm 0.5cm},width=30mm]{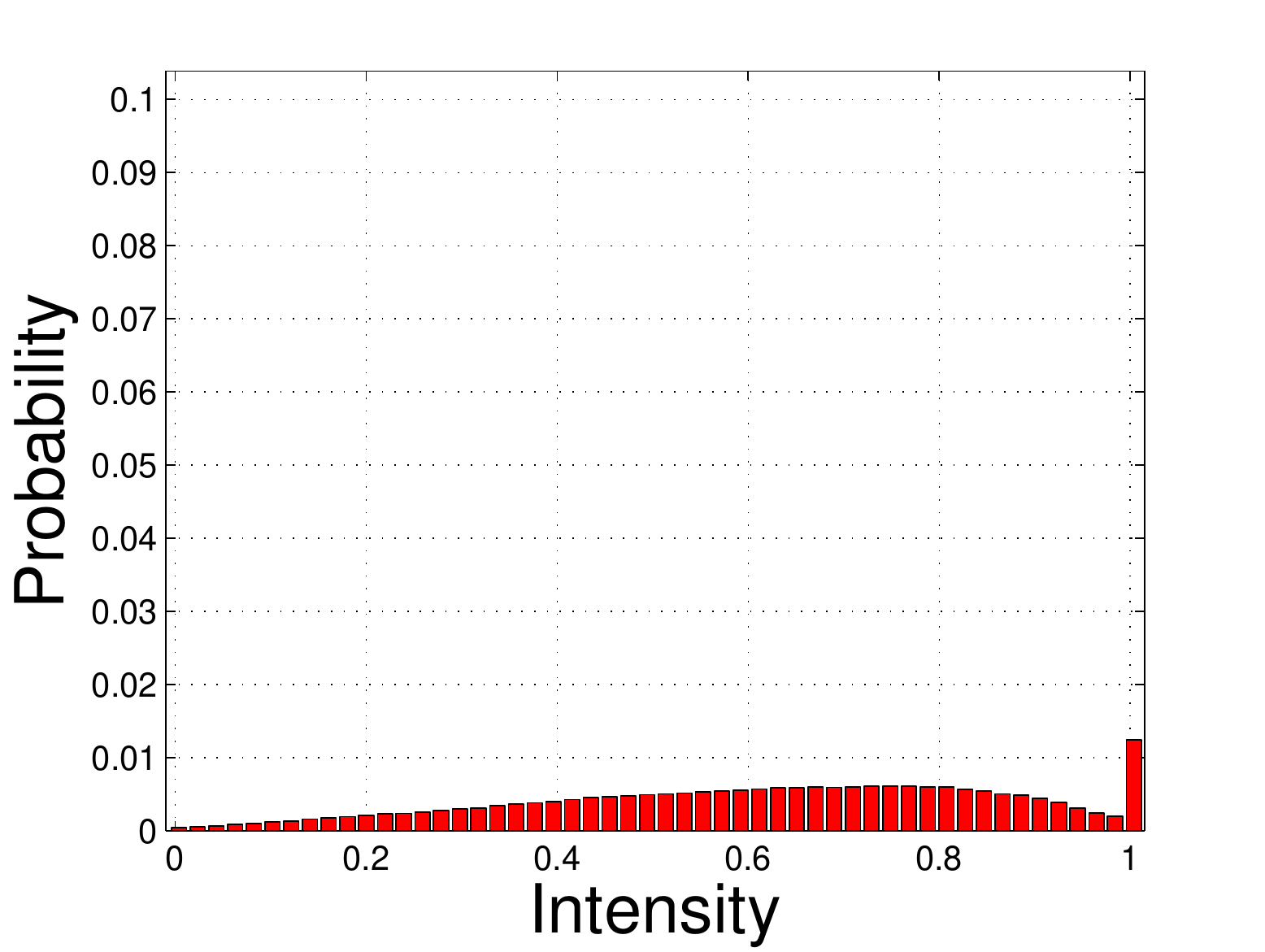}
\label{subfig:veil}
}
\subfloat[VDP \textit{P.M.}]{
\includegraphics[ trim={0cm 0.5cm 0cm 0.5cm},width=30mm]{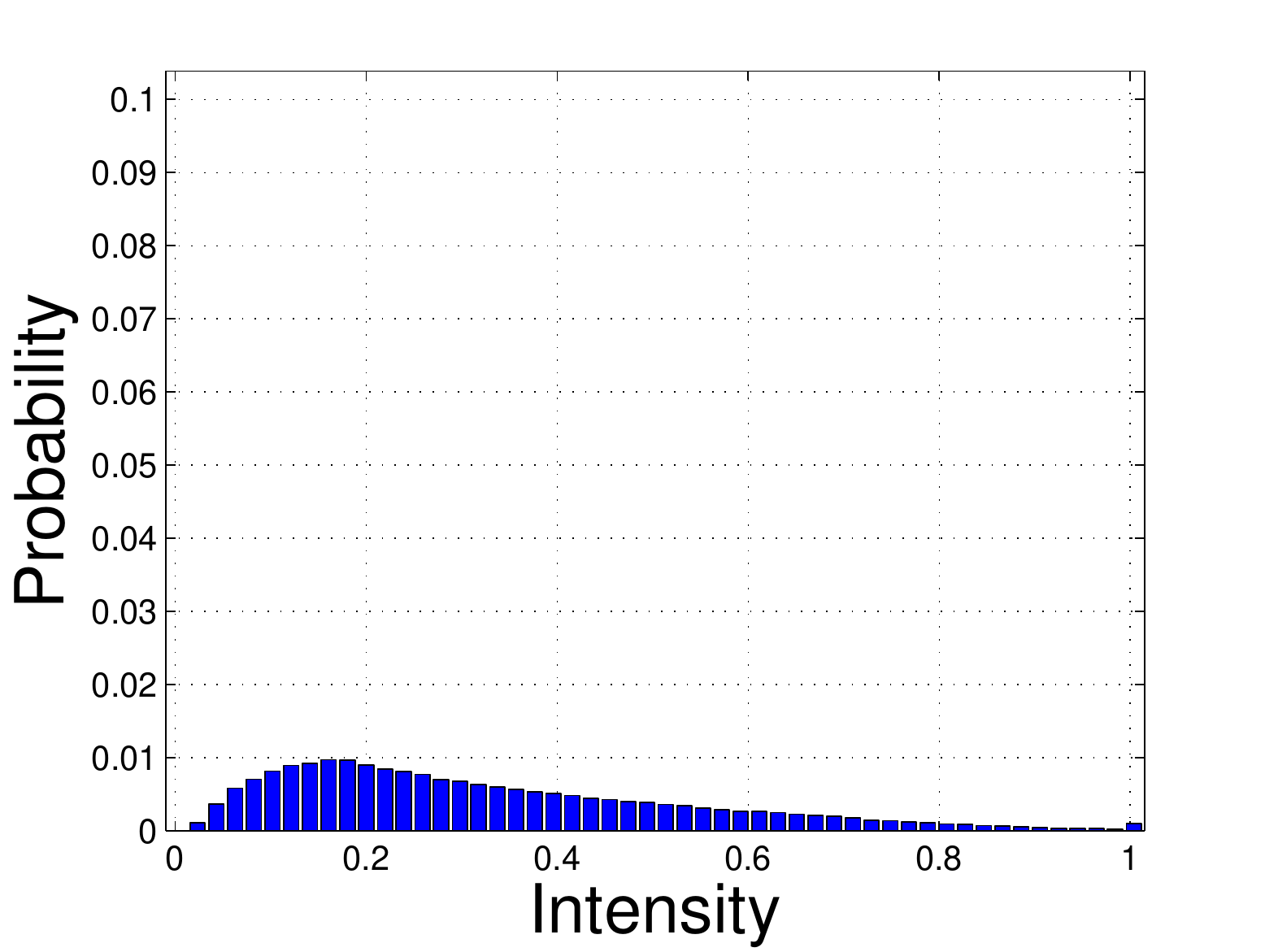}
\label{subfig:veilpm}
}\subfloat[Contrast \textit{H.F.}]{
\includegraphics[ trim={0cm 0.5cm 0cm 0.5cm},width=30mm]{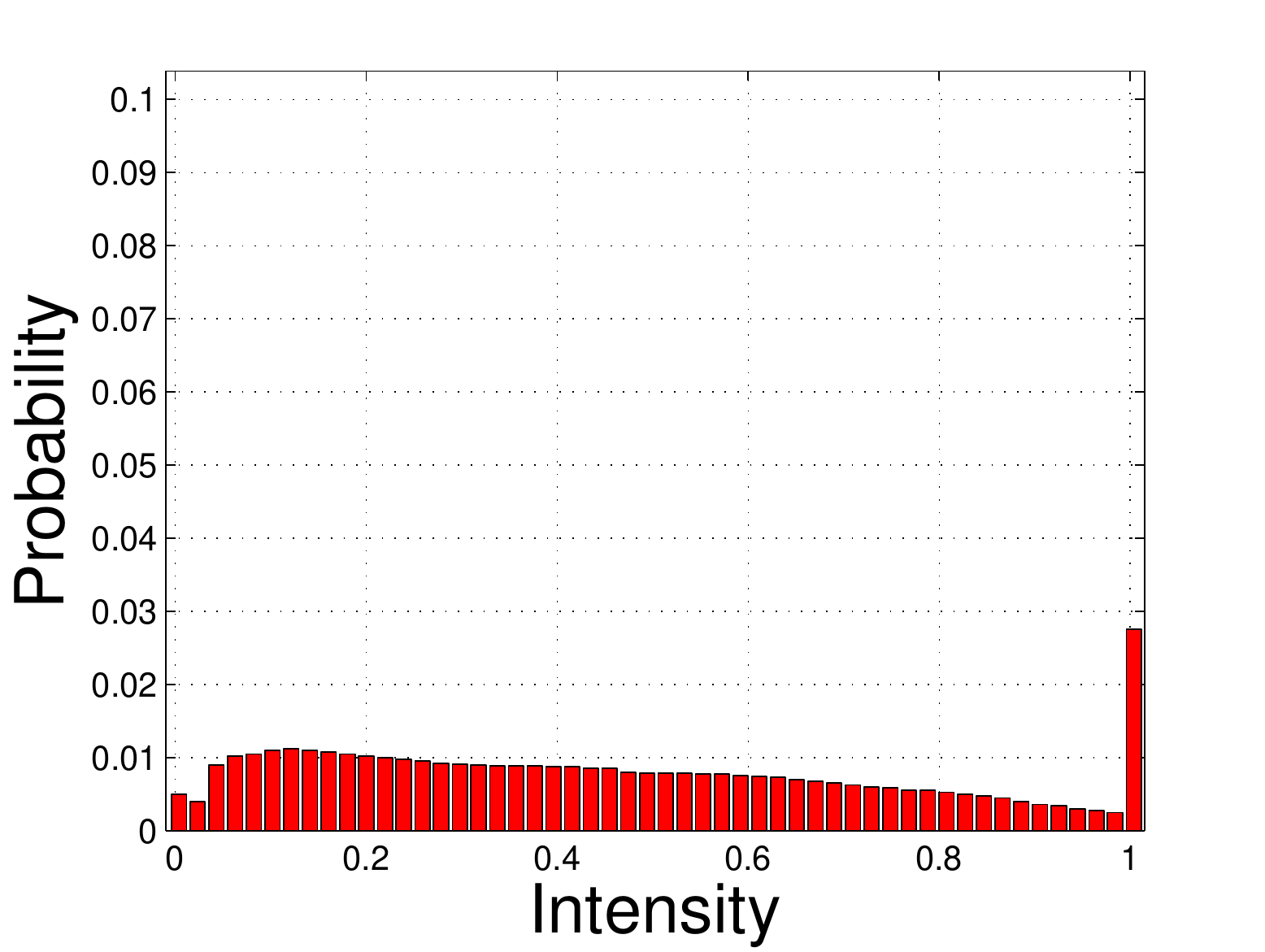}
\label{subfig:con}
}
\subfloat[Contrast \textit{P.M.}]{
\includegraphics[ trim={0cm 0.5cm 0cm 0.5cm},width=30mm]{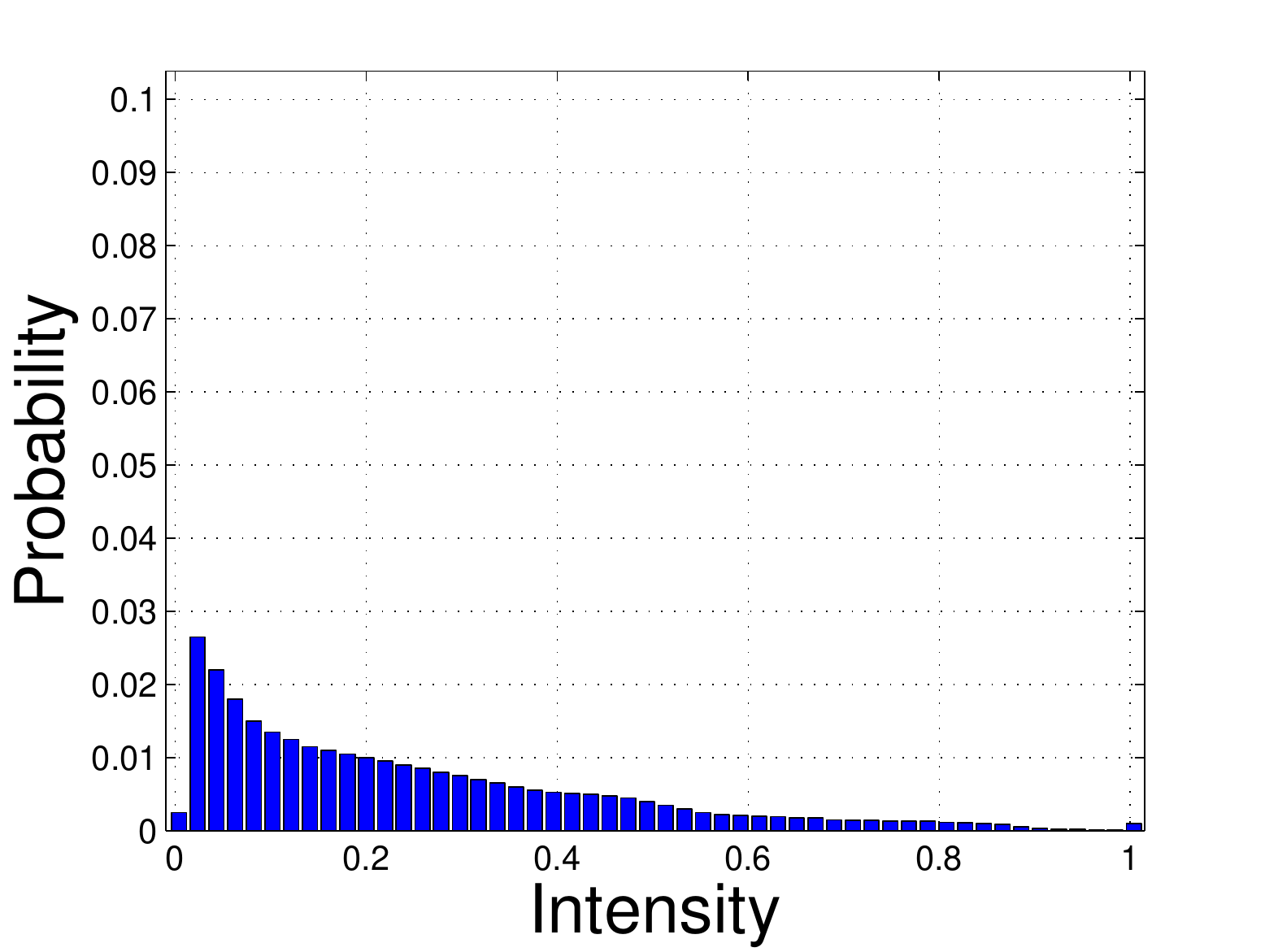}
\label{subfig:conpm}
}

\subfloat[Composite  \textit{H.F.}]{
\includegraphics[ trim={0cm 0.5cm 0cm 0.4cm},width=30mm]{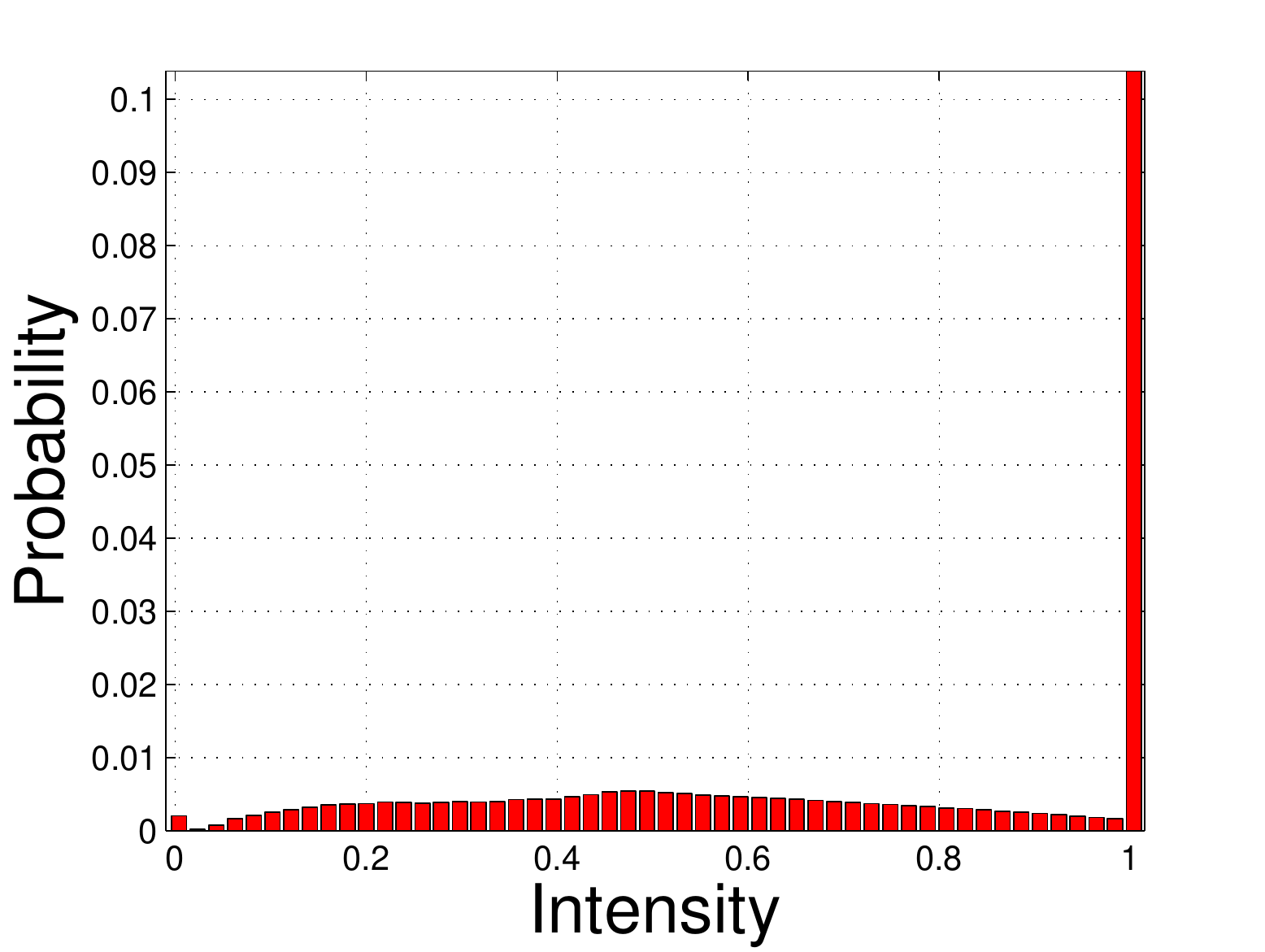}
\label{subfig:final}
}
\subfloat[Composite \textit{P.M.}]{
\includegraphics[ trim={0cm 0.5cm 0cm 0.4cm},width=33mm]{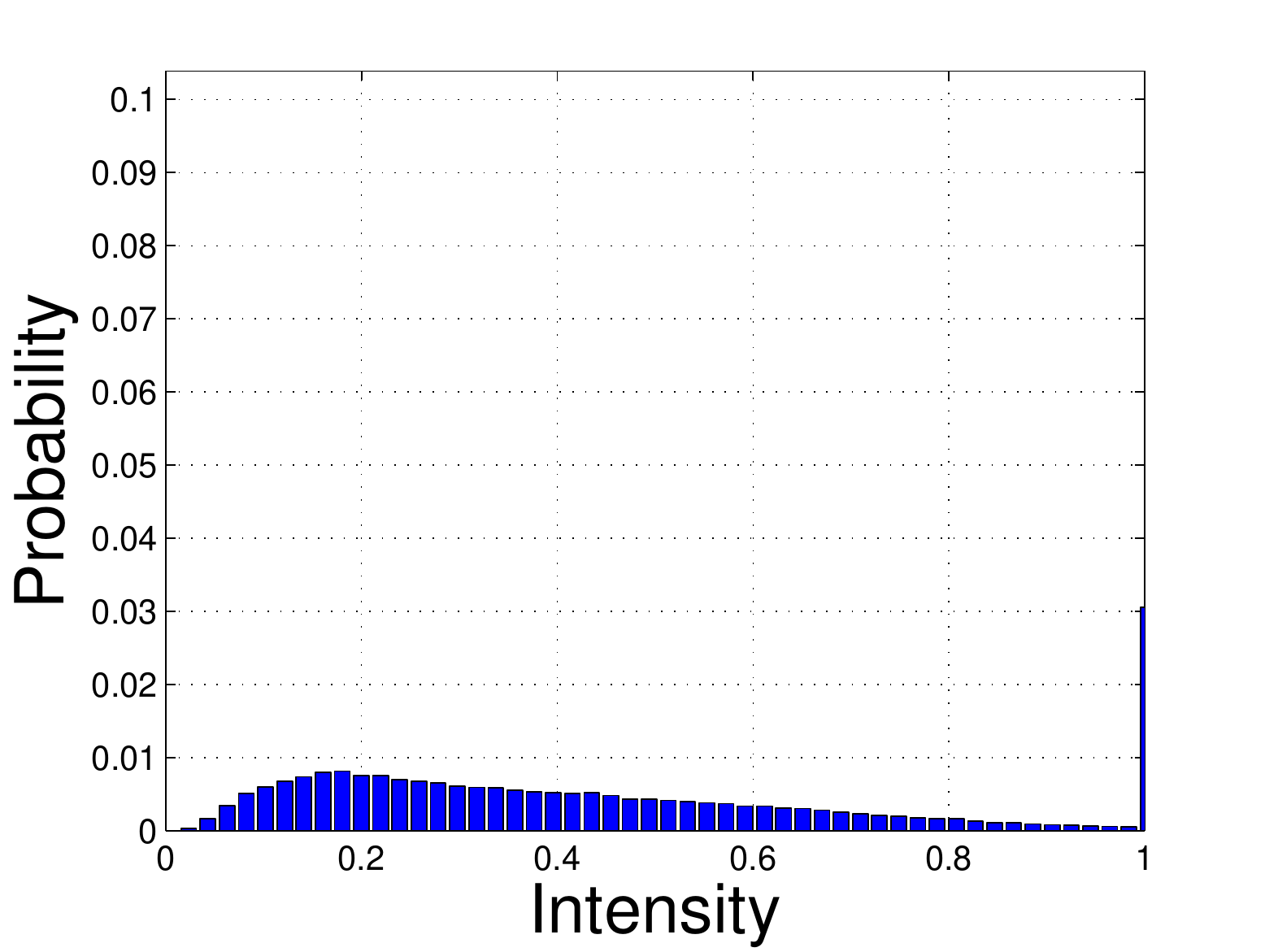}
\label{subfig:finalpm}
}
\caption{Average histograms for DCP \citep{He09}, UDCP \citep{Drews13}, Veil Difference Prior (VDP), Contrast Prior(CP) and Composite Prior.   
For haze-free (\textit{H.F.}) dataset we used 2,000 images from the test set of the popular ImageNet \citep{Imagenet15}, showed in red. For participating media (\textit{P.M.}), showed in blue, we built a dataset by collecting 2,000 images over the web. The images were resized  into a maximum side of 1,024 pixels and the priors were computed over patches of $15\times 15$ pixels.
For a better understanding of results,  we show sampling of the 256-bins histograms on every five bins.
Also for an easier comparison we plotted $1-DCP$ and also $1-UDCP$. In this evaluation, the target behavior is a high response for the haze-free dataset and a lower response for participating media dataset.
}\label{fig:histo}
\end{figure}

This function could be assumed as combination of the transmissions. In this paper, we assume a conservative function. We believe that, when a pixel present a high transmission value, this indication is usually related to the information of the signal, and unrelated to the veiling. Thus, we propose a simple combination of transmission estimators by using a max function. We denote the transmission of a pixel $x$ as:

\begin{equation}
 t(x) = \max(t_v(x),t_c(x)), 
\label{eq:trans}
\end{equation}
\noindent where $t_v(x)$ is the refined transmission
computed with the \textit{Veil Difference Prior} and
$t_c(x)$ is the refined transmission computed with the \textit{Contrast Prior}.
We also show the statistical analysis for the Composite Transmission, arguably to be the most reliable behavior. For haze-free images (Fig. \ref{subfig:final}) there is a clear tendency of assuming the value one than only the veil difference prior. It combines the best of two worlds, having high range of values for participating media images, and a solid high response for haze-free images. This corroborates to obtain the generality of the proposed method.

We also show the contribution from each of the image priors after applying Eq. (\ref{eq:trans}) on Fig. \ref{subfig:contribution}.
We perceive that the intermediate range transmissions are dominated by the contrast transmission. The final transmission is show on Fig. \ref{subfig:matting}.

%---------------------------------------------------------------------

\subsection{Restoration}
\label{sec:rest}

\begin{figure}[!ht]
\centering
\subfloat[Initial]{
\includegraphics[width=27mm]{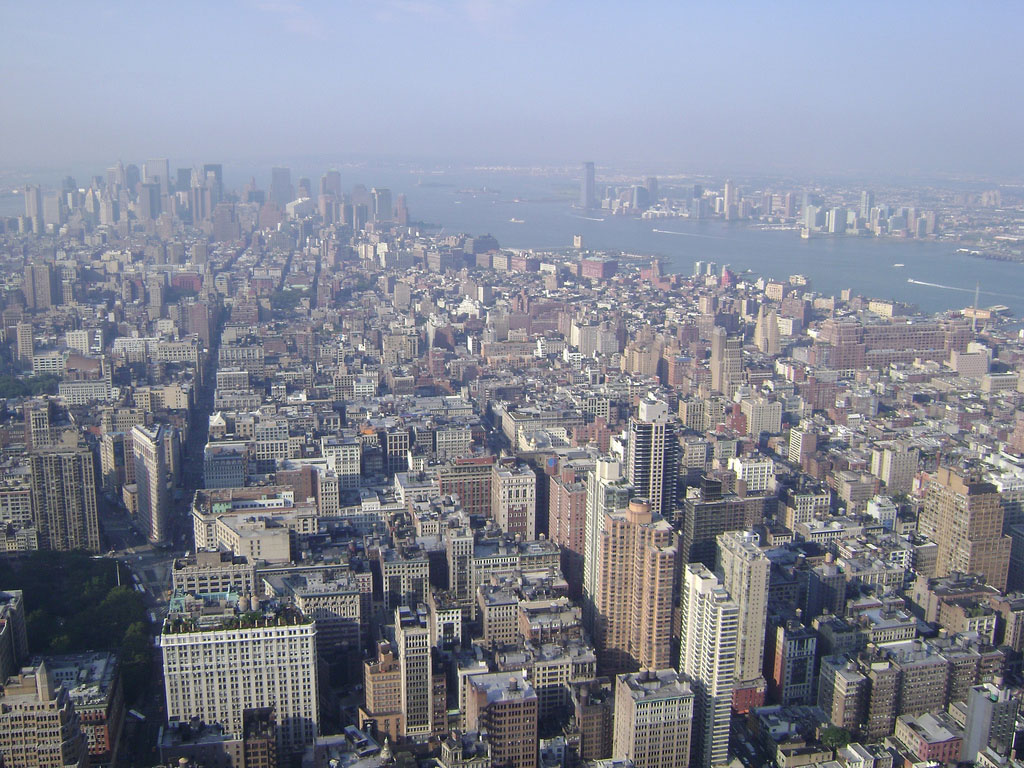}
}
\subfloat[$\tilde{t}_v(x)$]{
\includegraphics[width=27mm]{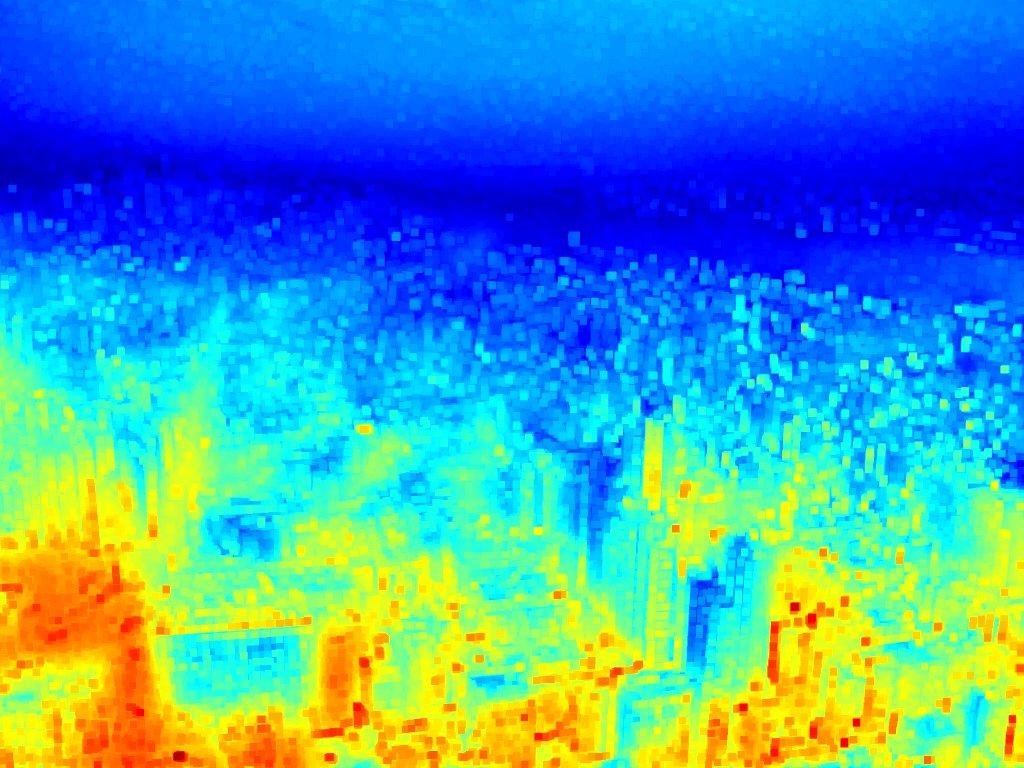}
\label{subfig:tv}
}
\subfloat[$\tilde{t}_c(x)$]{
\includegraphics[width=27mm]{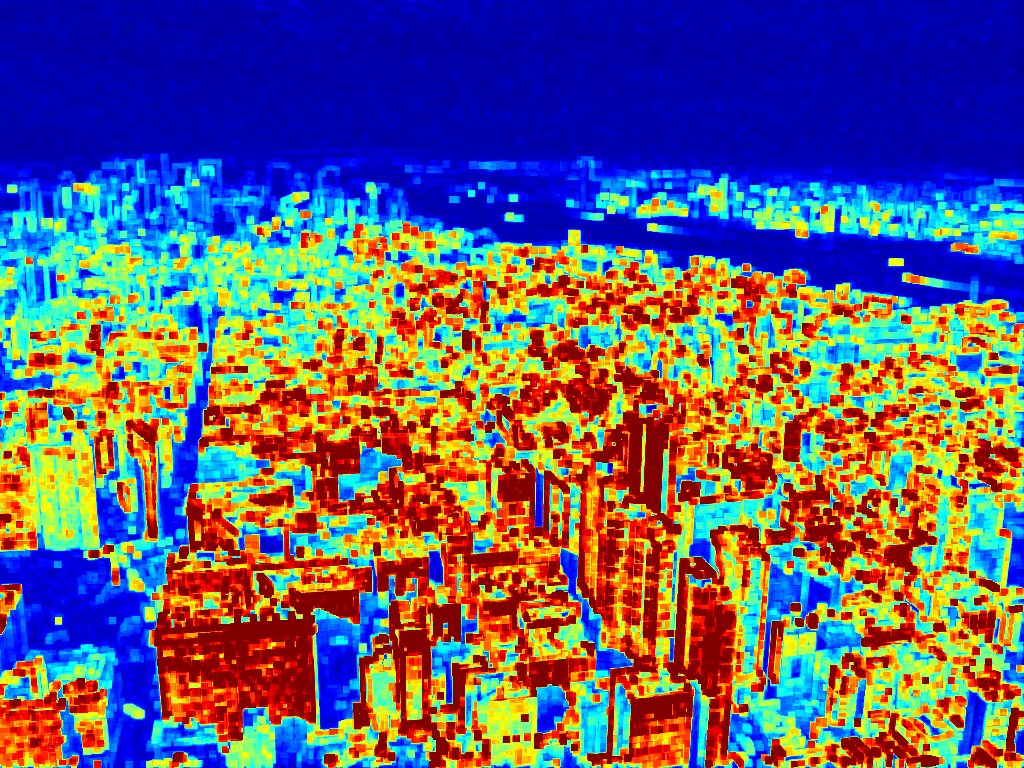}
\label{subfig:tc}
}
\subfloat[$t_v(x)$]{
\includegraphics[width=27mm]{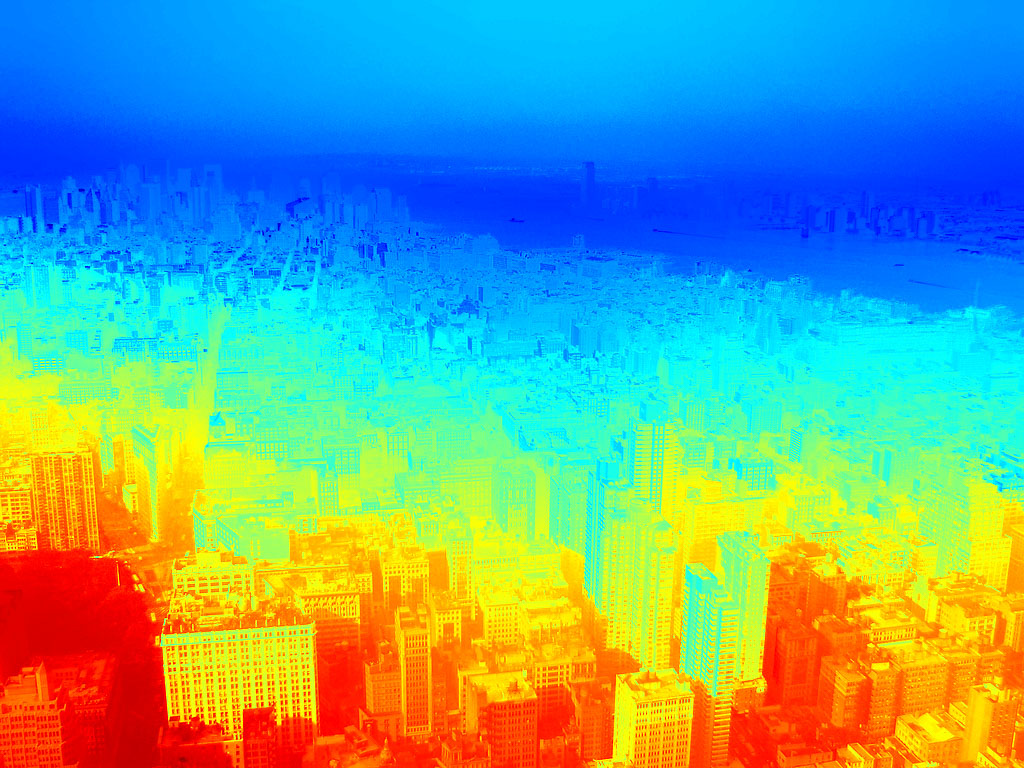}
\label{subfig:tvref}
}

\subfloat[$t_c(x)$]{
\includegraphics[width=27mm]{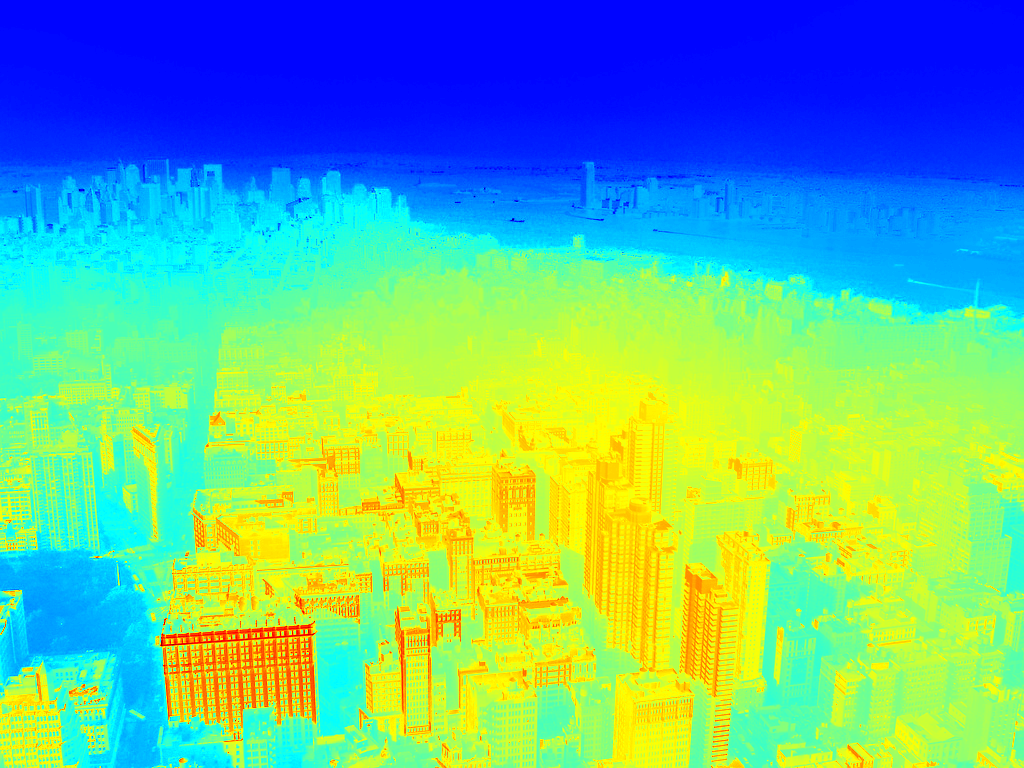}
\label{subfig:tcref}
}
\subfloat[Contributions]{
\includegraphics[width=27mm]{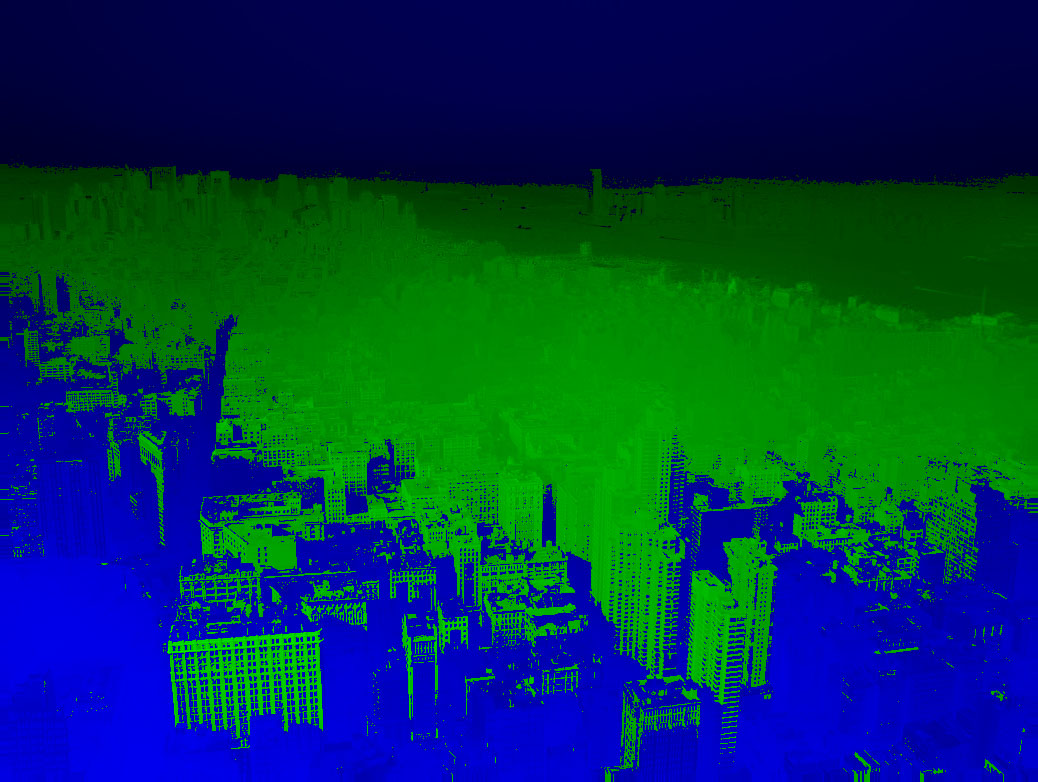}
\label{subfig:contribution}
}
\subfloat[Final]{
\includegraphics[width=27mm]{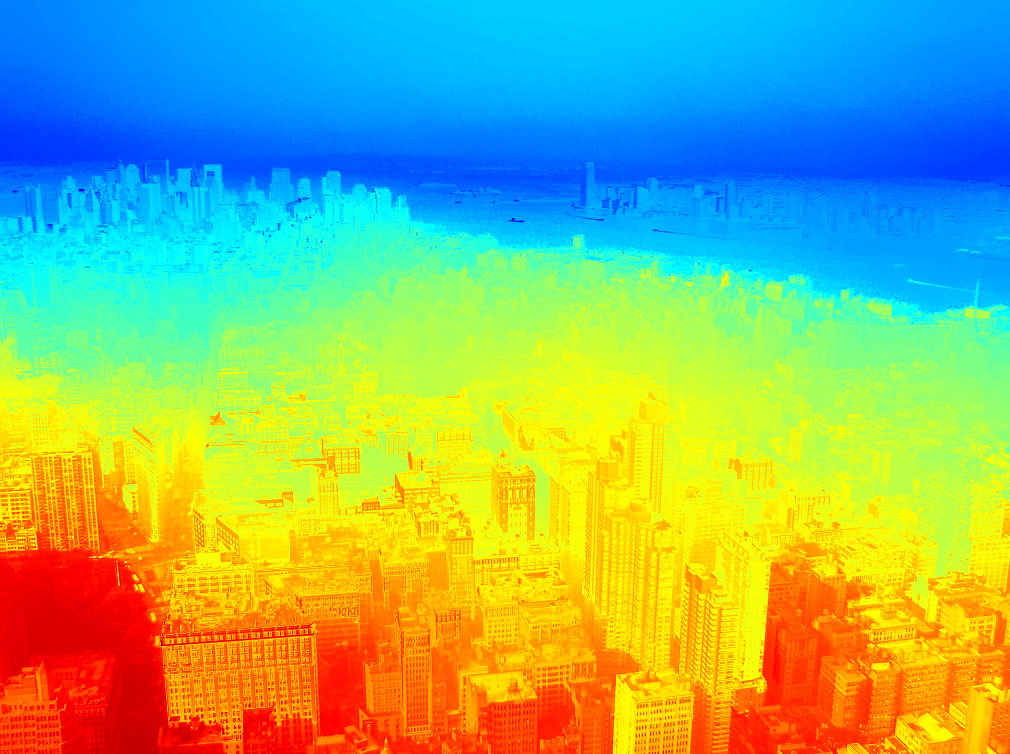}
\label{subfig:matting}
}
\subfloat[Restored Image]{
\includegraphics[width=27mm]{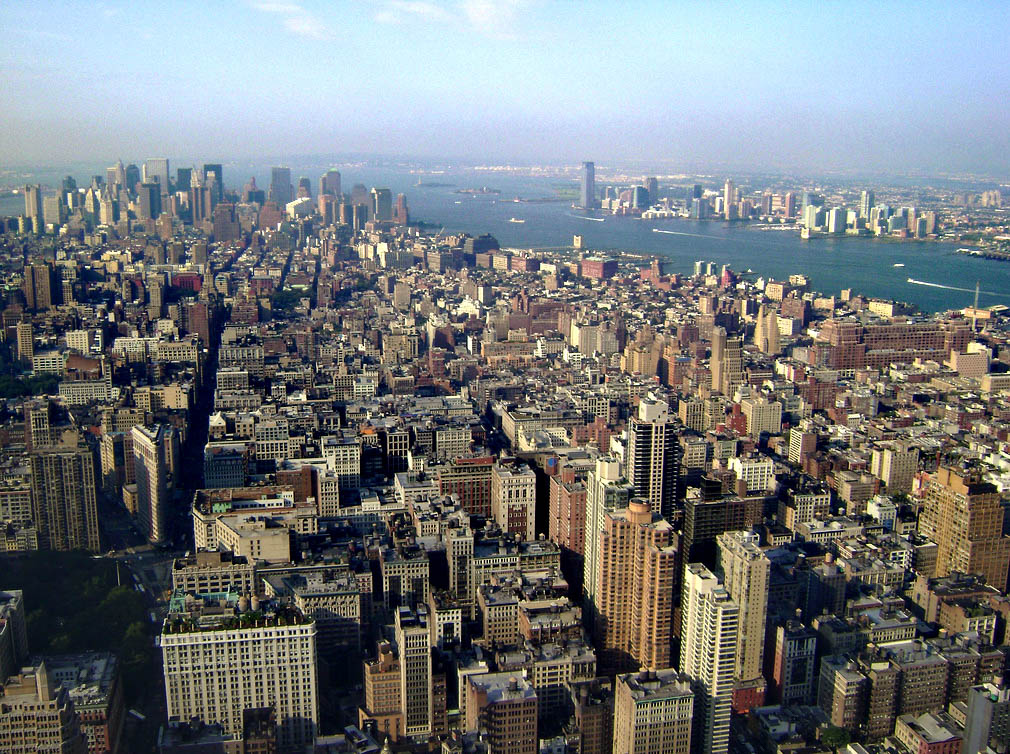}
\label{subfig:rest}
}
\caption{The result obtained by the proposed method and the transmission maps:
(a) The input fog image; (b) Veil difference
transmission $\tilde{t}_v(x)$; (c) Contrast transmission $\tilde{t}_c(x)$; (d) Refined veil difference transmission $t_v(x)$;(e) Refined contrast transmission $t_c(x)$;(f) The contribution
of each transmission after the $\max$ operator, where green is $t_c(x)$ and blue
is $t_v(x)$; (g) The final transmission;
(h) The restored image.}
\label{fig:compare}
\end{figure}

The image with restored visibility (see Fig. \ref{subfig:rest}) is obtained
by isolating the object reflectivity $M_\lambda(x)$ on Eq. \ref{eq:final} \footnote{We assume $A_\lambda^D$ can be estimated to compute the transmission and the object reflectivity.}:

\begin{equation}
 M_\lambda(x) = \frac{I_\lambda(x) - A_\lambda^D + A_\lambda^Dt(x)}{\max(t_{\lambda_0},A_\lambda^Dt(x))}, \label{eq:rest}
\end{equation}
\noindent where the $t_{\lambda_0}$ parameter is the minimum transmission. This parameter avoids to restore noise when there are no information behind the veil. Typical values range between $[0.1, 0.2]$. 

%---------------------------------------------------------------------

\subsection{Estimating the Ambient Light}

Several authors estimate ambient light as the brightest pixels in the image \citep{Tan08,Fattal08}. This estimation presents some drawbacks, specially if the ambient light is not present on the image. \cite{Sulami14} proposed a method that divides the estimation into orientation and magnitude and does not need the presence of an ambient light pixel on the image. However, it depends on finding patches that obeys certain properties.

As we stated on Sec. \ref{sec:model}, the ambient light is associated with the light source on the scene. Thus, it is reasonable to use color constancy techniques, commonly adopted to find the light source color on general images. Techniques
such as gray edge \citep{Van07}, the gray world, the max-rgb or the shades-of-gray \citep{Finlayson04} can be used. The problem of the gray edge technique is the edges are normally blurry and the gray world technique fails when there are too many close objects. Max-rgb is dependent on  white patches fully reflecting the light information. We find the shades-of-gray algorithm encapsulates the best behavior for participating medium. The algorithm obtains an estimation in between the average of the scene and the reflectivity of a white patch.
% Close Composite Transmission Estimation
%---------------------------------------------------------------------

%-------------------------------------------------------------------------
% Experimental Evaluation
%
\section{Experimental Evaluation}
\label{sec:evaluation}

The experimental results are obtained by a standard C++ with OpenCV library. All the parameters are defined as explained on Sec \ref{sec:rest}. We compared the results with state-of-the-art methods. For most of the cases, we adopted the provided images by the authors
in order to reproduce their results. The Red Channel \citep{Galdran15} and DCP \citep{He09} transmissions were implemented according to the paper by using also a standard C++ with OpenCV. The UDCP \citep{Drews13,Drews16} results  are obtained using the implementation provided by the authors.

%---------------------------------------------------------------------
\subsection{Qualitative Evaluation}

\begin{figure}[!ht]
\centering
\subfloat[Input]{
\includegraphics[width=28mm]{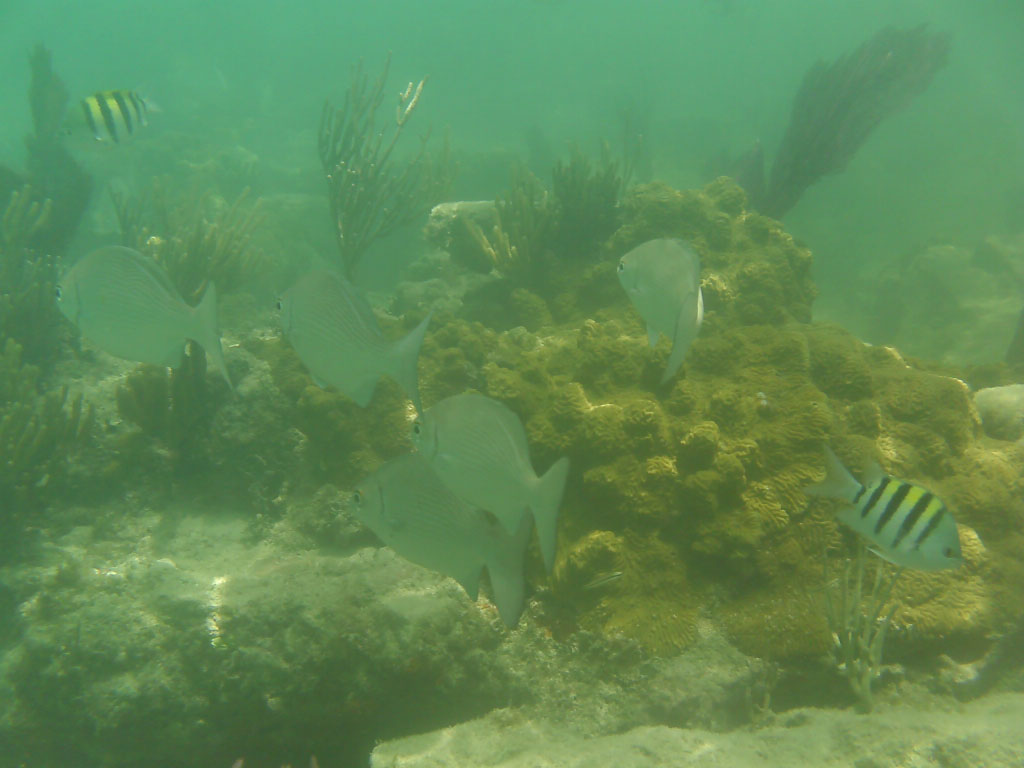}
\label{subfig:a3}
}\subfloat[\cite{Ancuti12}]{
\includegraphics[width=28mm]{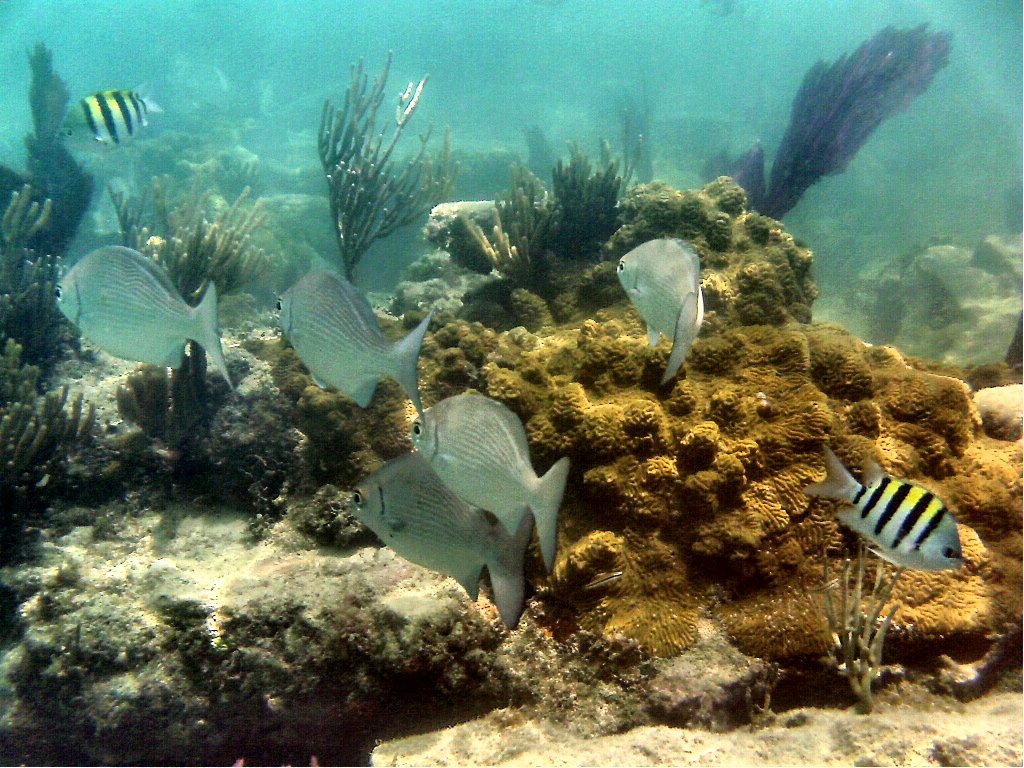}
\label{subfig:a3an}
}\subfloat[Input]{
\includegraphics[width=28mm]{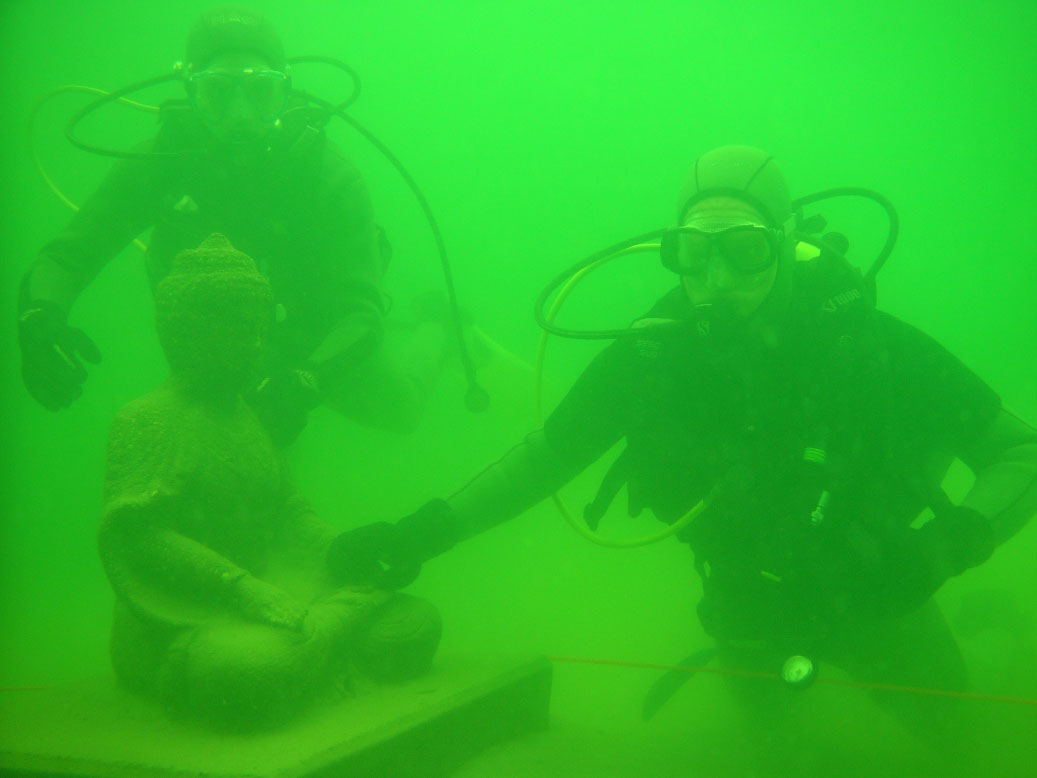}
\label{subfig:a6}
}\subfloat[\cite{Ancuti12}]{
\includegraphics[width=28mm]{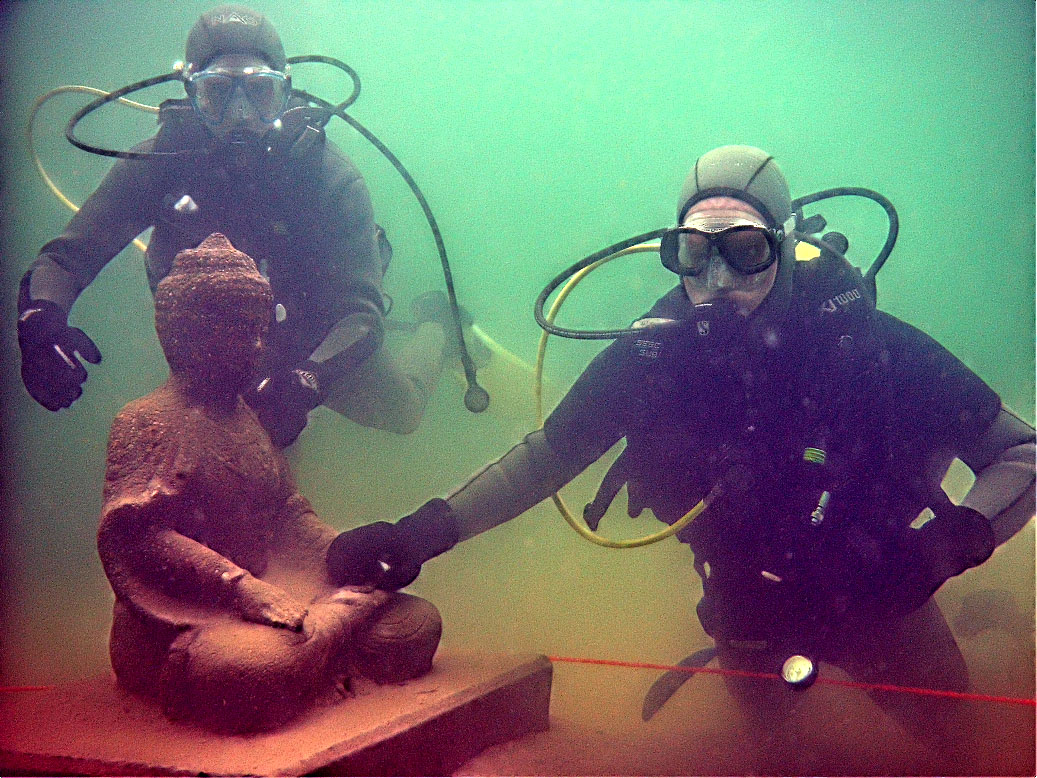}
\label{subfig:a6an}
}

\subfloat[\cite{Drews13}]{
\includegraphics[width=28mm]{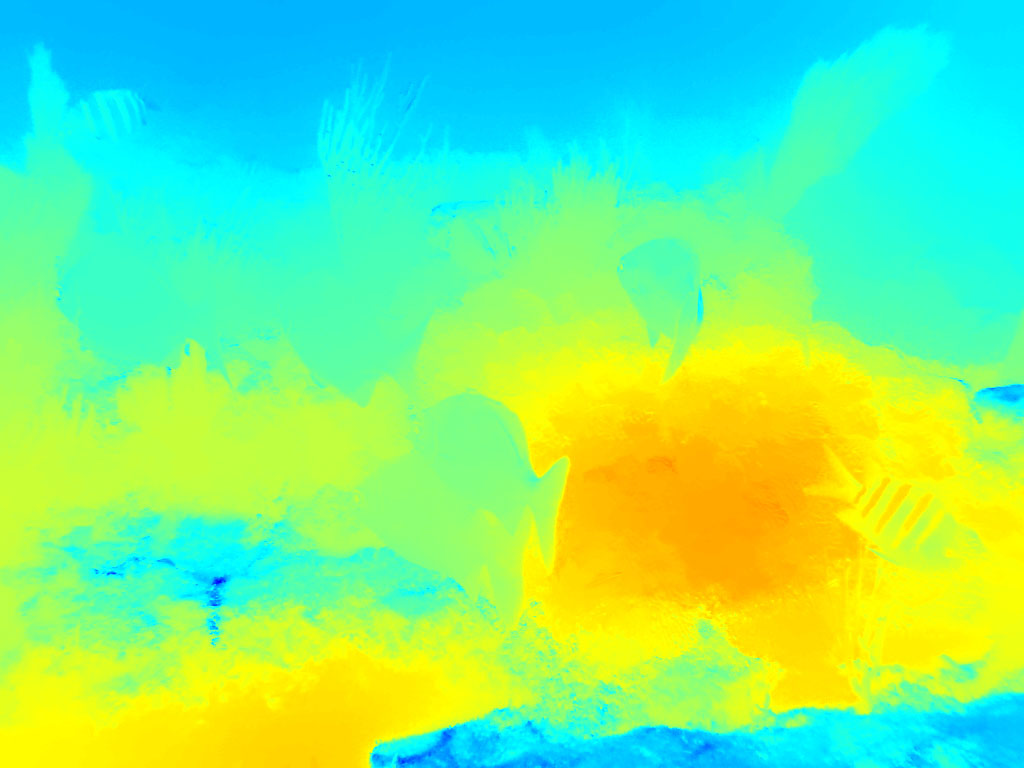}
\label{subfig:a3the}
}\subfloat[\cite{Drews13}]{
\includegraphics[width=28mm]{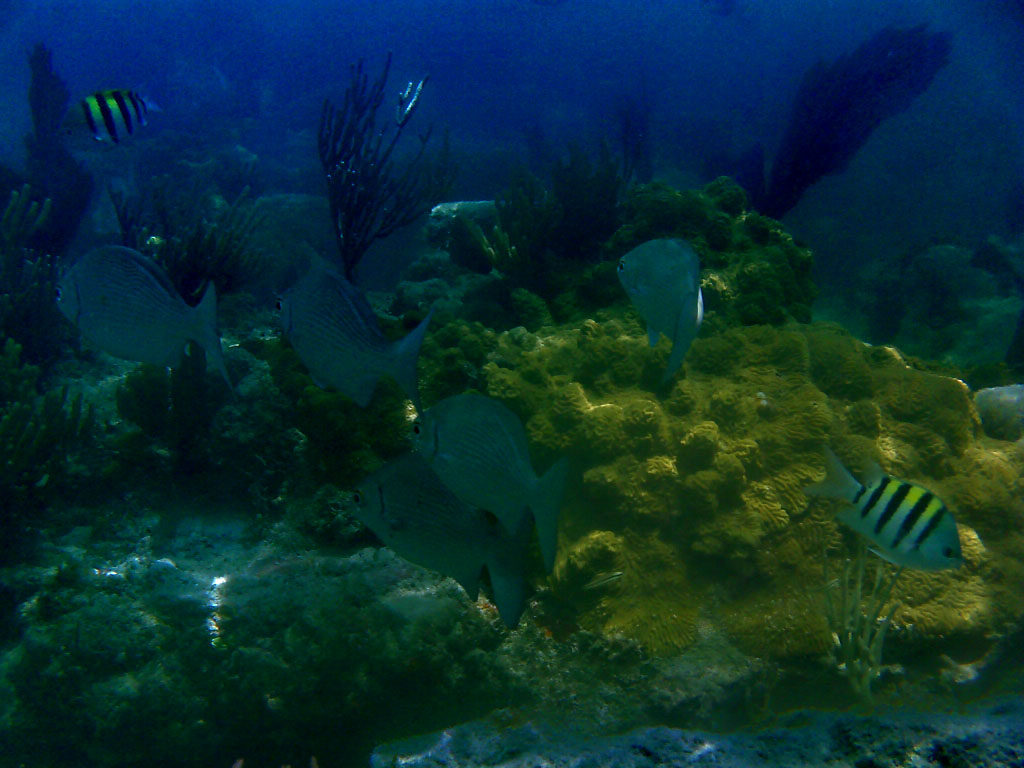}
\label{subfig:a3he}
}\subfloat[\cite{Drews13}]{
\includegraphics[width=28mm]{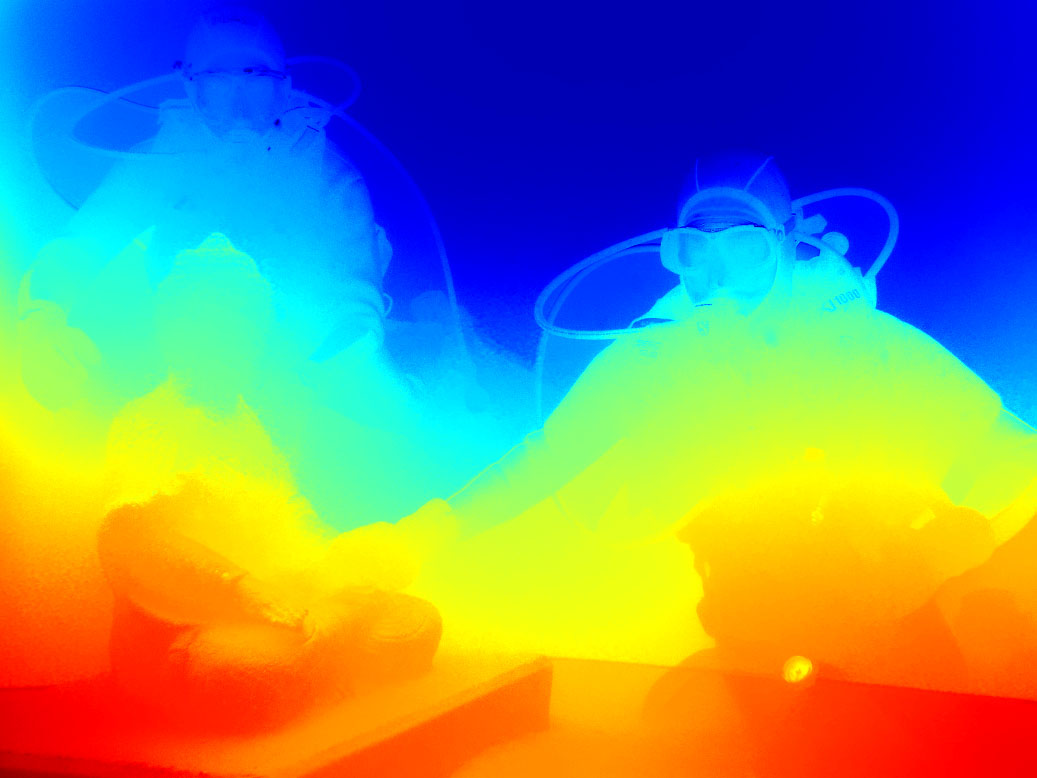}
\label{subfig:3f12}
}\subfloat[\cite{Drews13}]{
\includegraphics[width=28mm]{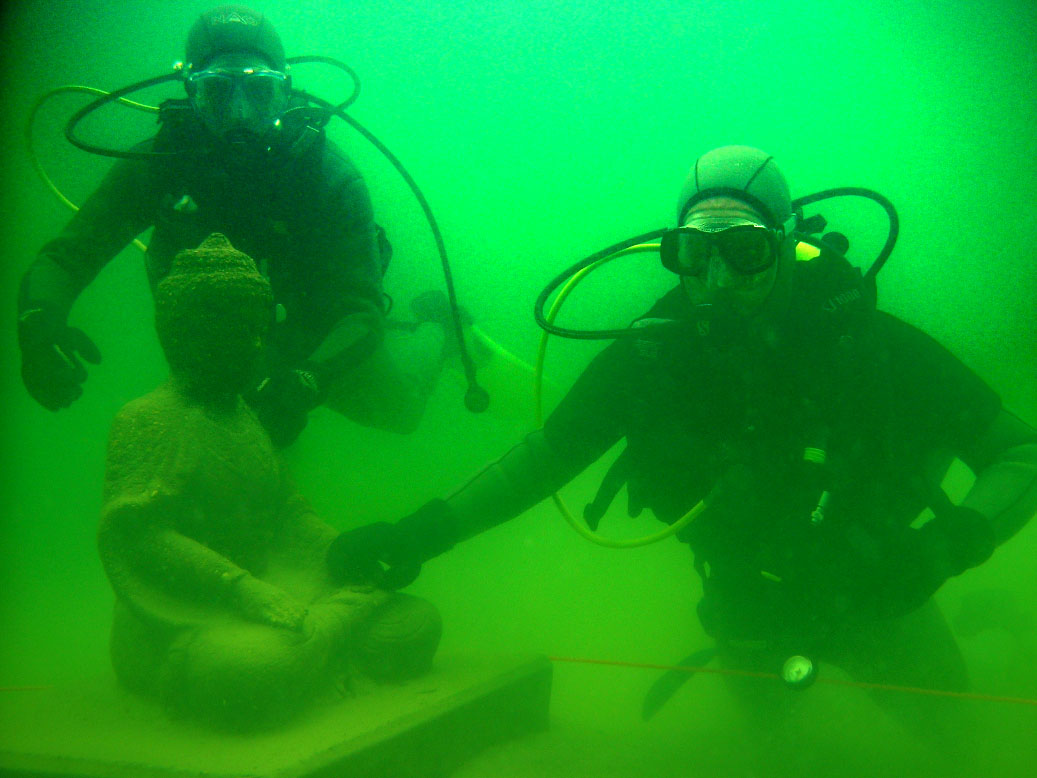}
\label{subfig:1f23}
}

\subfloat[\cite{Galdran15}]{
\includegraphics[width=28mm]{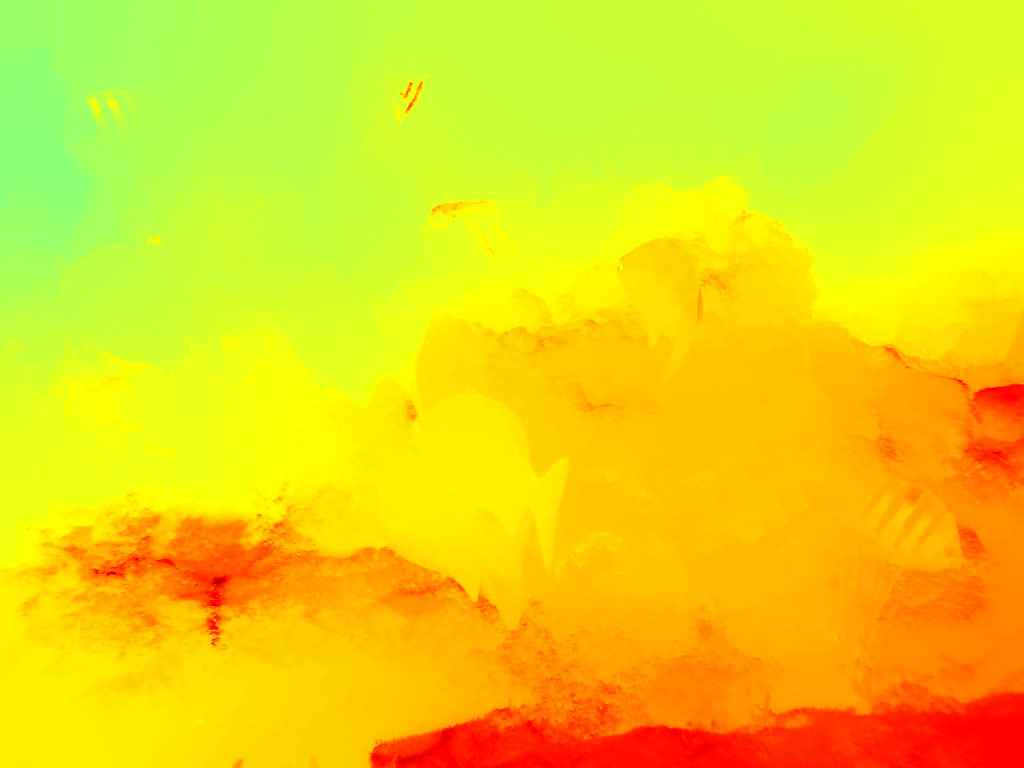}
\label{subfig:3f10}
}\subfloat[\cite{Galdran15}]{
\includegraphics[width=28mm]{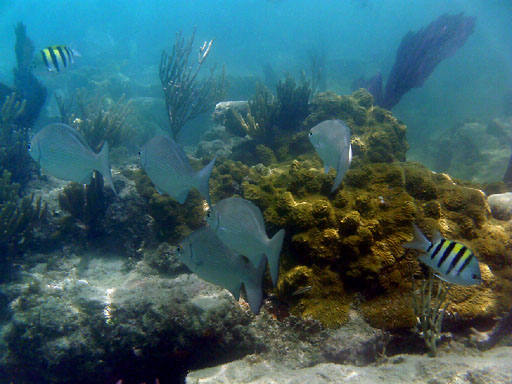}
\label{subfig:3f20}
}\subfloat[\cite{Galdran15}]{
\includegraphics[width=28mm]{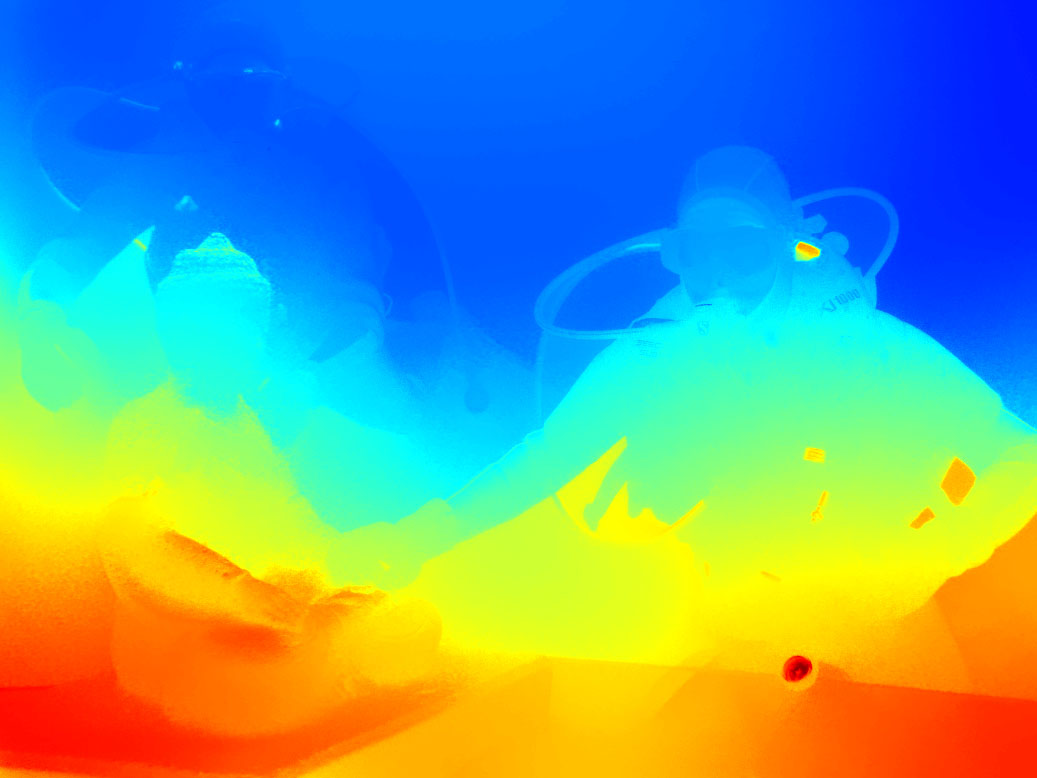}
\label{subfig:3f18}
}\subfloat[\cite{Galdran15}]{
\includegraphics[width=28mm]{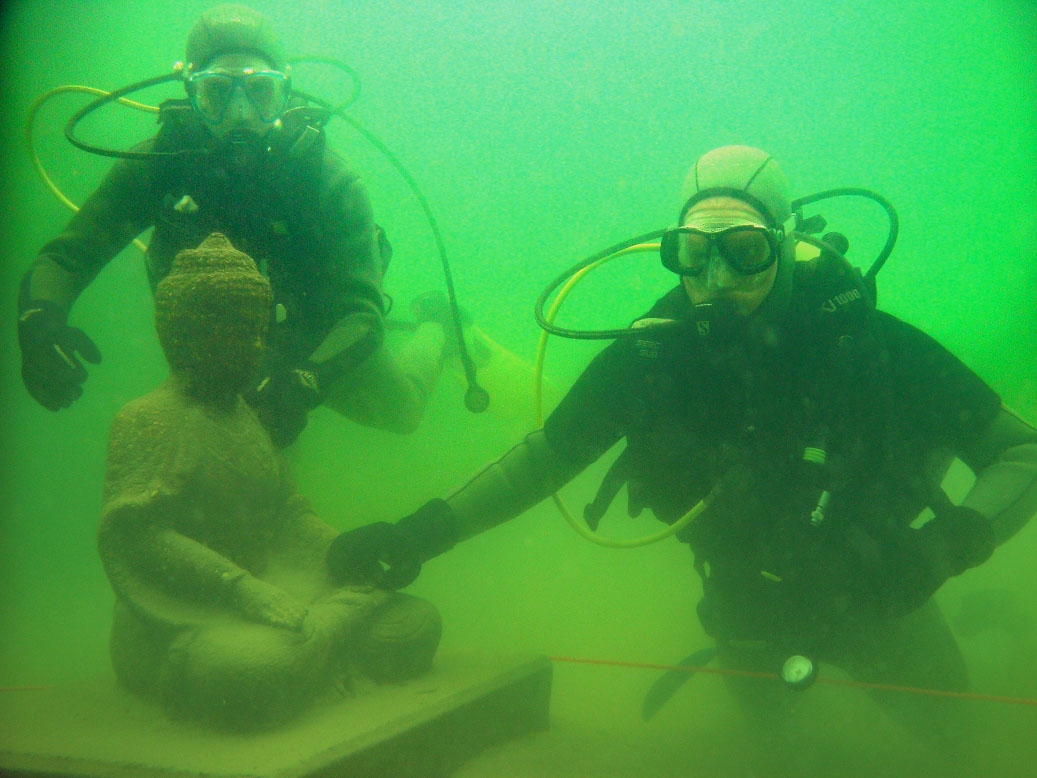}
\label{subfig:3f28}
}

\subfloat[Our Transmission]{
\includegraphics[width=28mm]{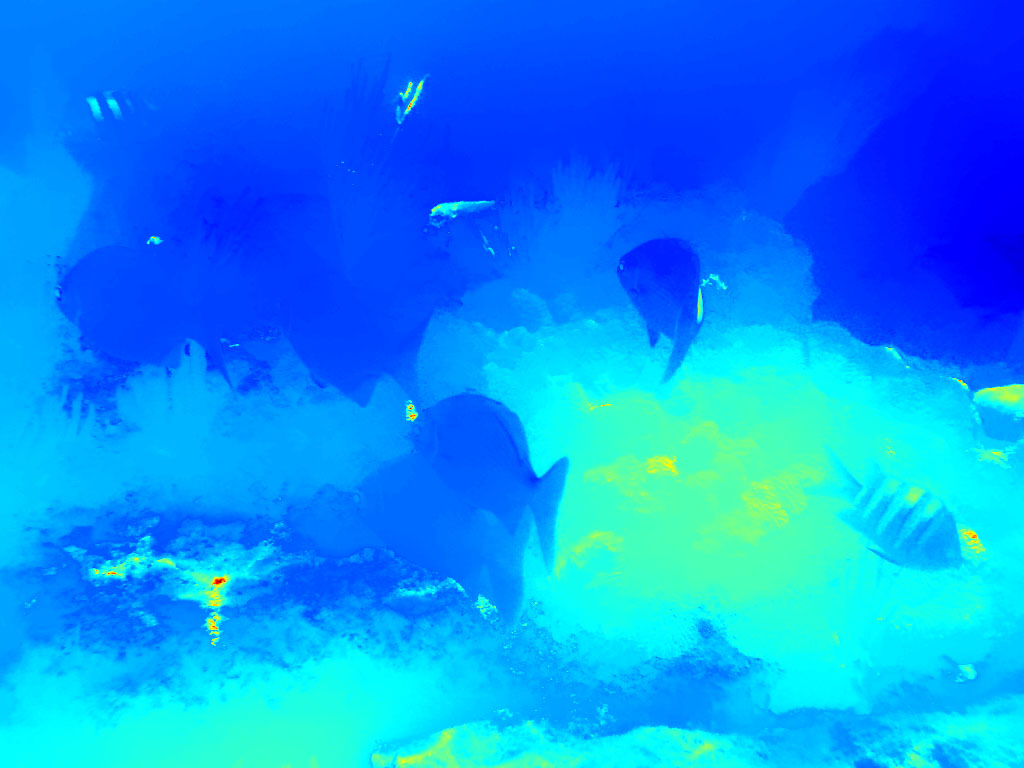}
\label{subfig:a3t}
}\subfloat[Our Restoration]{
\includegraphics[width=28mm]{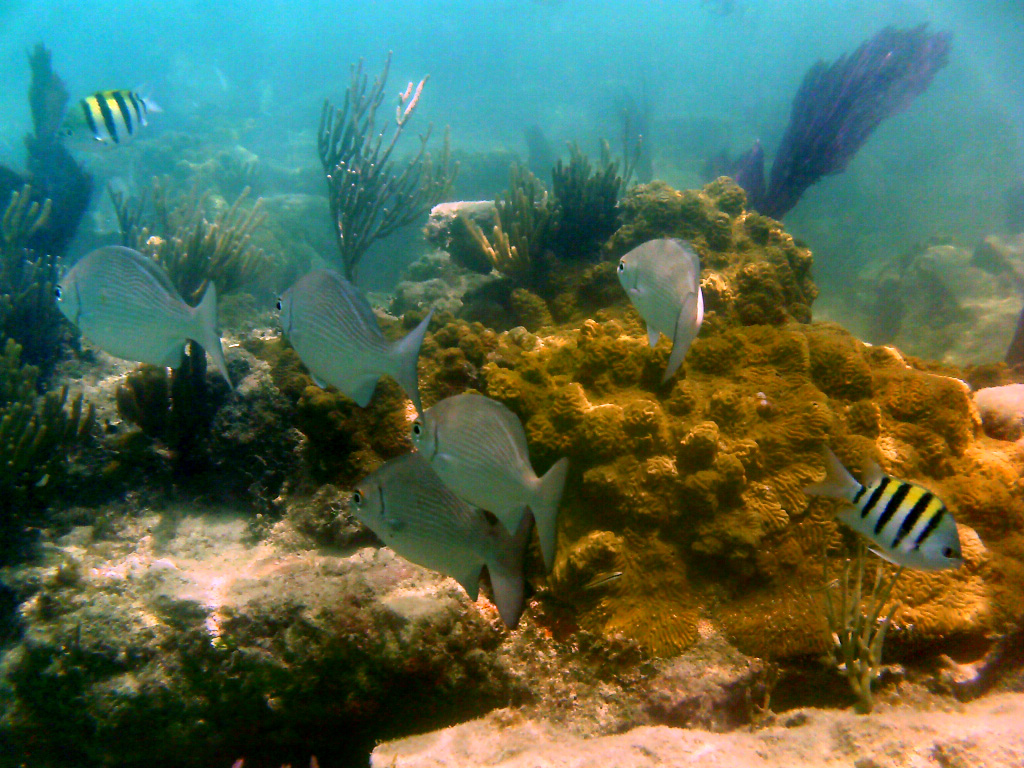}
\label{subfig:a3ours}
}\subfloat[Our Transmission]{
\includegraphics[width=28mm]{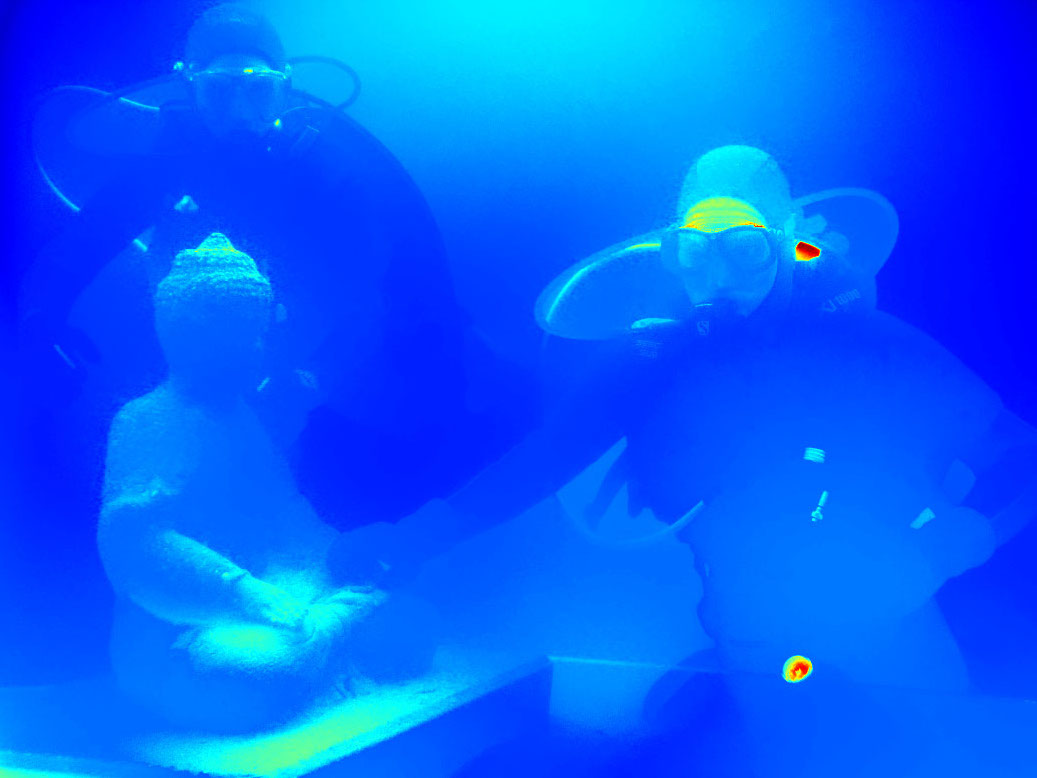}
\label{subfig:a6tours}
}\subfloat[Our Restoration]{
\includegraphics[width=28mm]{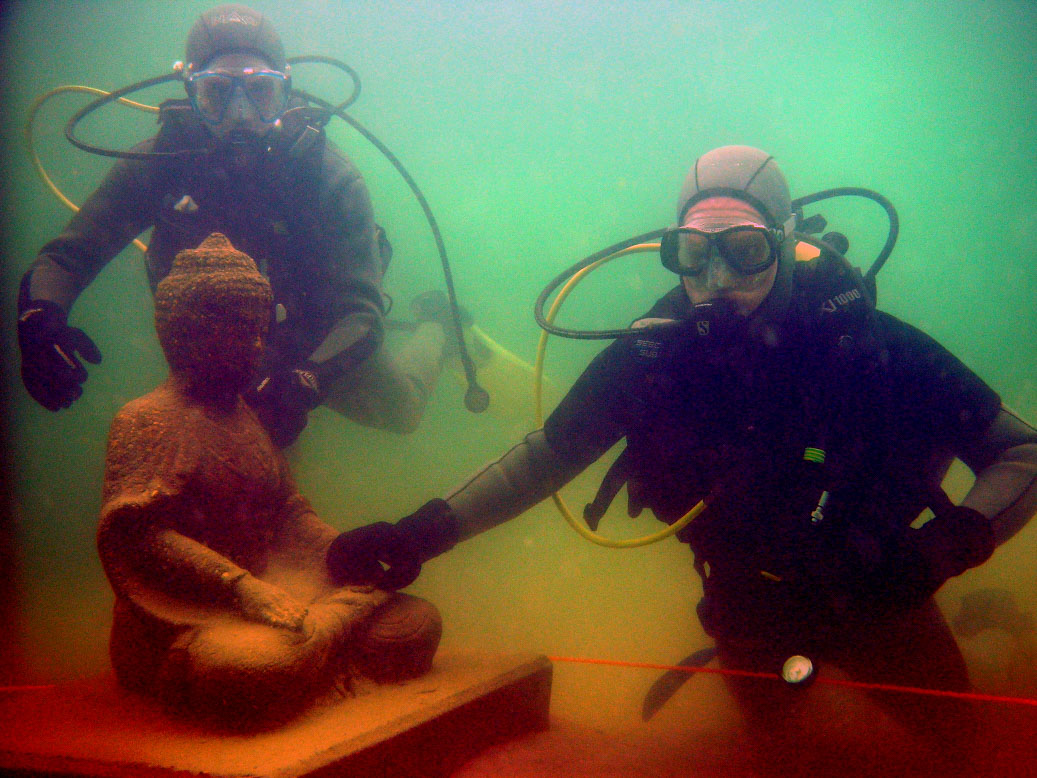}
\label{subfig:0f38}
}

\caption{Results showing the transmission estimation and restoration for underwater images. The transmissions are shown using a red to blue scale, where red indicates a higher transmission. The obtained results are compared with \cite{Galdran15}, \cite{Drews13} and \cite{Ancuti12}. The images are obtained from \cite{Ancuti12}. }
\label{fig:underwater}
\end{figure}

Fig. \ref{fig:underwater} shows the results obtained for the underwater environment. The proposed method tends to not overestimate transmission. This results is mainly due to the observation that the ambient light color is not necessarily present on the image. 
The Red Channel \citep{Galdran15} assumption works quite well for Fig. \ref{subfig:a3} with a small overestimation. However, red channel method clearly presents problems by overestimating transmission on red objects with high amount of degradation (\textit{e.g.} Fig. \ref{subfig:3f18}).

There is a clear tendency of our method to present more contrasted colors and less noise, but with slightly less contrast when comparing the results with \cite{Ancuti12} (Figs. \ref{subfig:a3an} and \ref{subfig:a6an}). It is due to the use of optical model of the participating media. Our method is capable of recovering color properties without further use of white balances and compensations. This can be seen when comparing the proposed method with the UDCP \citep{Drews13}, where a good restoration is obtained (Fig \ref{subfig:a3he}) but with an unsatisfactory color correction. 

\begin{figure}[!htb]
\centering
\subfloat{
\includegraphics[width=15mm]{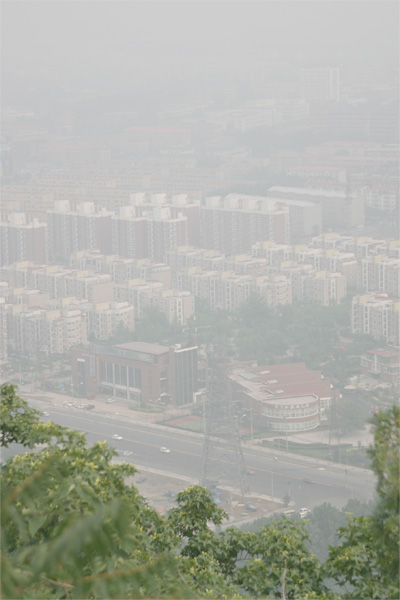}
}
\subfloat{
\includegraphics[width=15mm]{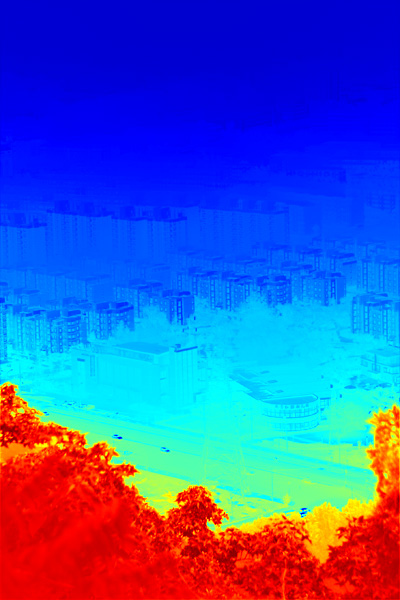}
\label{subfig:2f1_}
}
\subfloat{
\includegraphics[width=15mm]{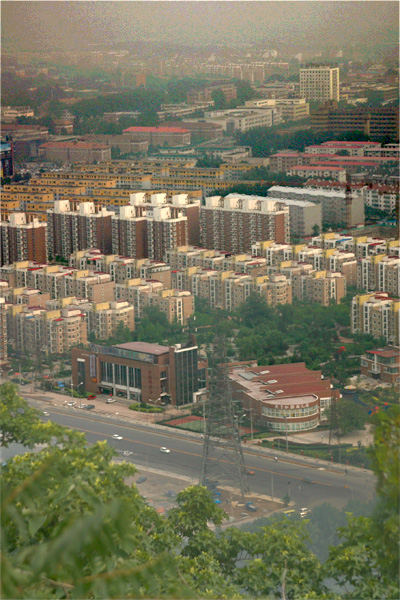}
\label{subfig:2f2_}
}
\subfloat{
\includegraphics[width=15mm]{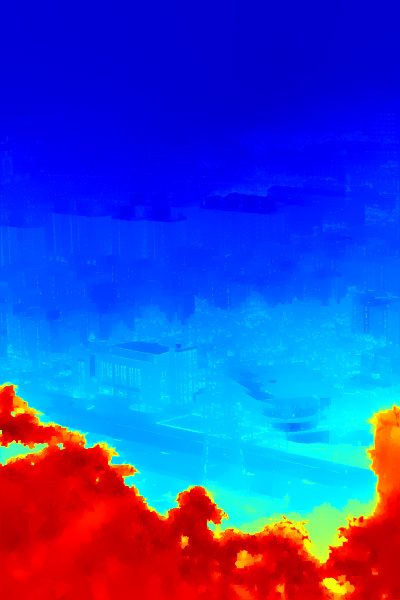}
\label{subfig:3f1_}
}
\subfloat{
\includegraphics[width=15mm]{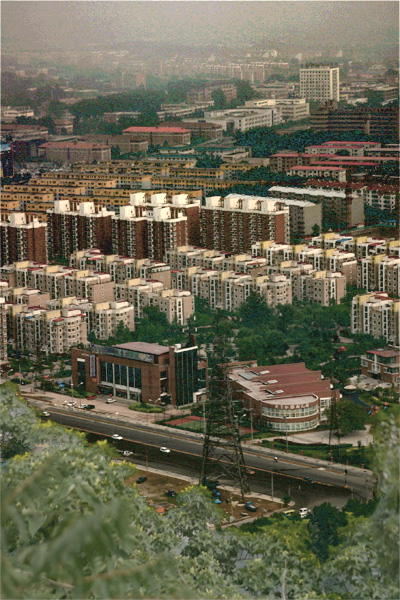}
\label{subfig:3f2_}
}
\subfloat{
\includegraphics[width=15mm]{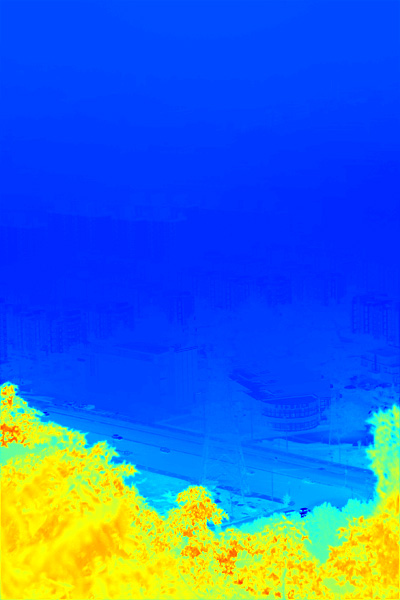}
\label{subfig:ourtrans_}
}
\subfloat{
\includegraphics[width=15mm]{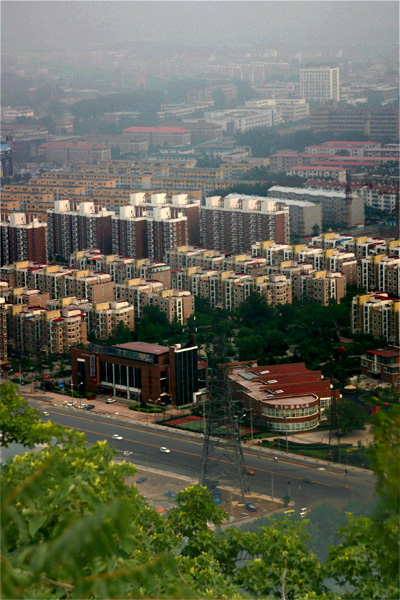}
\label{subfig:our_}
}
\caption{Results for a haze image restoration: (a) Input image;
(b) Transmission from the \textit{DCP} \citep{He09}; (c)
Restoration from the \textit{DCP} \citep{He09}; (d) Transmission from \cite{Fattal14};
(e) Restoration from \cite{Fattal14}; (f) Our transmission; (g) Our restoration.}
\label{fig:haze}
\end{figure}

As stated earlier, our method is designed to restore any kind of participating media without any parameter adjustment. On Fig. \ref{fig:haze} we show the results on a haze image comparing with other state-of-the-art methods. We suppose the ambient light is not a pixel from the scene, thus it  helps to avoid the over-estimation of the transmission on lower distances. However, due the max operation between the priors, on longer distances, our method culminates into overestimating the transmission. Finally, we also see a better color correction when compared to \cite{Fattal14} and DCP\citep{He09}. We explain this color enhancement due to the assumption that the haze is not perfectly white. 
%---------------------------------------------------------------------

\subsection{Quantitative Evaluation}

The literature usually compares the methods using a qualitative perception of image quality. However, it is hard to evaluate which is the best restored image, \textit{e.g} Figs. \ref{subfig:a3an} and \ref{subfig:a3ours}.

\begin{figure}[!htb]
\centering
\subfloat[Clean Image (T0)]{
\includegraphics[width=30mm]{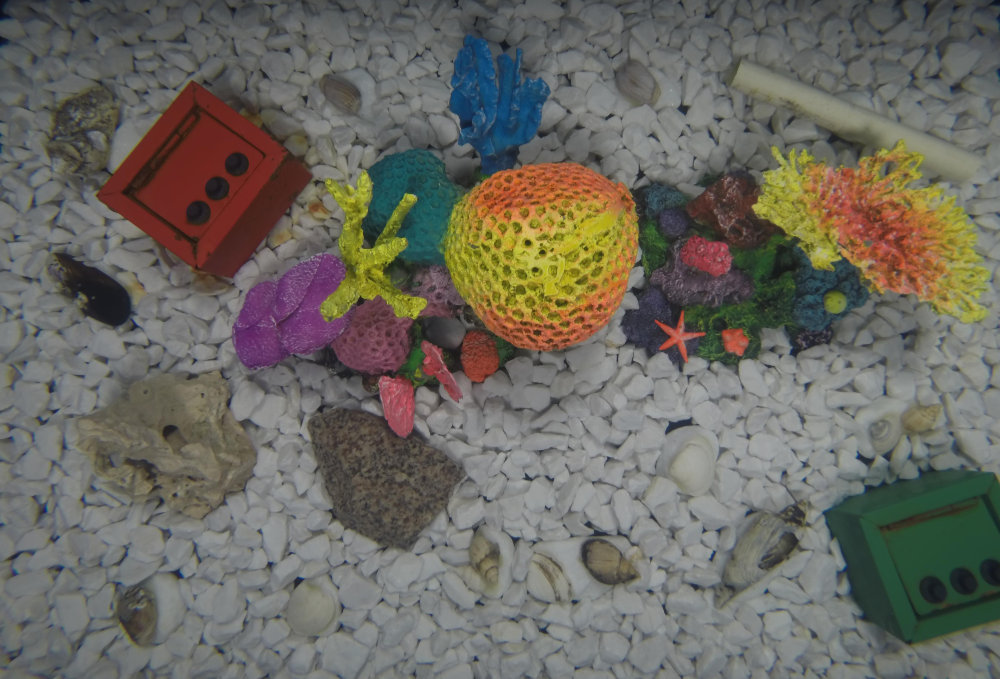}
\label{subfig:clean}
}
\subfloat[20ml (T5)]{
\includegraphics[width=30mm]{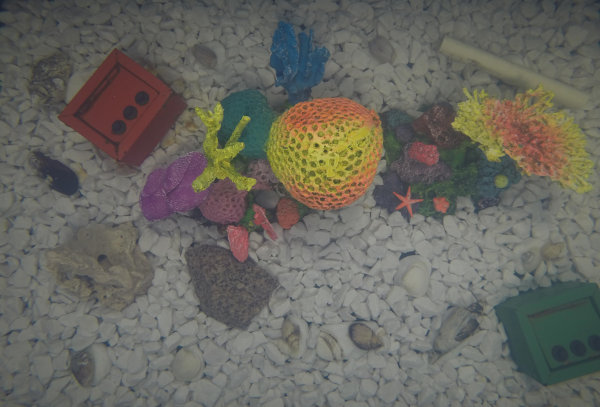}
\label{subfig:3f2}
}
\subfloat[58ml  (T11)]{
\includegraphics[width=30mm]{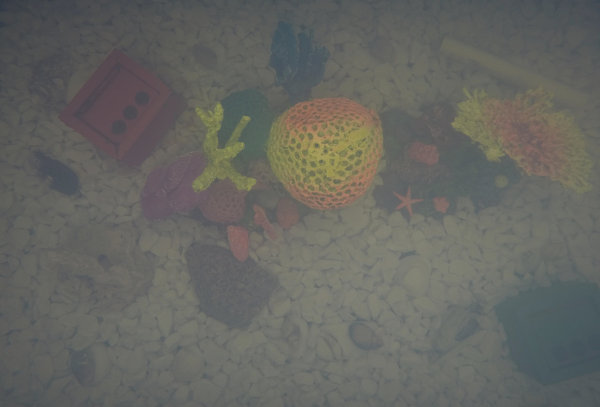}
\label{subfig:0f3}
}
\subfloat[110ml (T20)]{
\includegraphics[width=30mm]{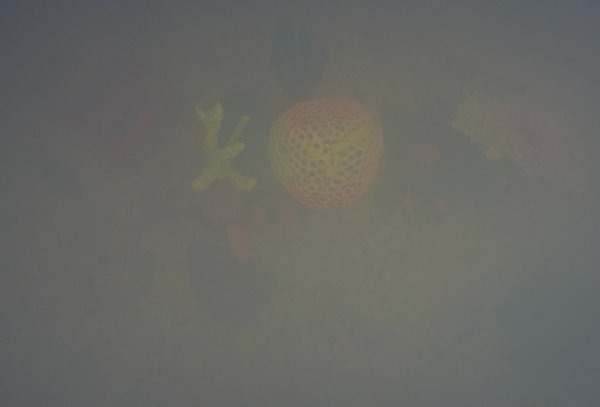}
\label{subfig:0f4}
}
\caption{Samples of the TURBID dataset collected by \cite{duarte2016dataset}.}
\label{fig:turbid}
\end{figure}

An effective way to evaluate dehazing algorithms quality is by comparing them with a ground truth image, \textit{i.e.} a version of the image without the effects of the medium. For this, we adopted the TURBID dataset collected by \citep{duarte2016dataset}. They reproduced a scene of an underwater environment where multiple images are captured with an increasing degradation by the addition of milk. Fig. \ref{fig:turbid} shows four image samples of the TURBID dataset. 

%--------------------------------------------------------------

\begin{figure}[!h]
\centering
\subfloat[$A_\lambda^D$ fixed]{
\includegraphics[width=60mm]{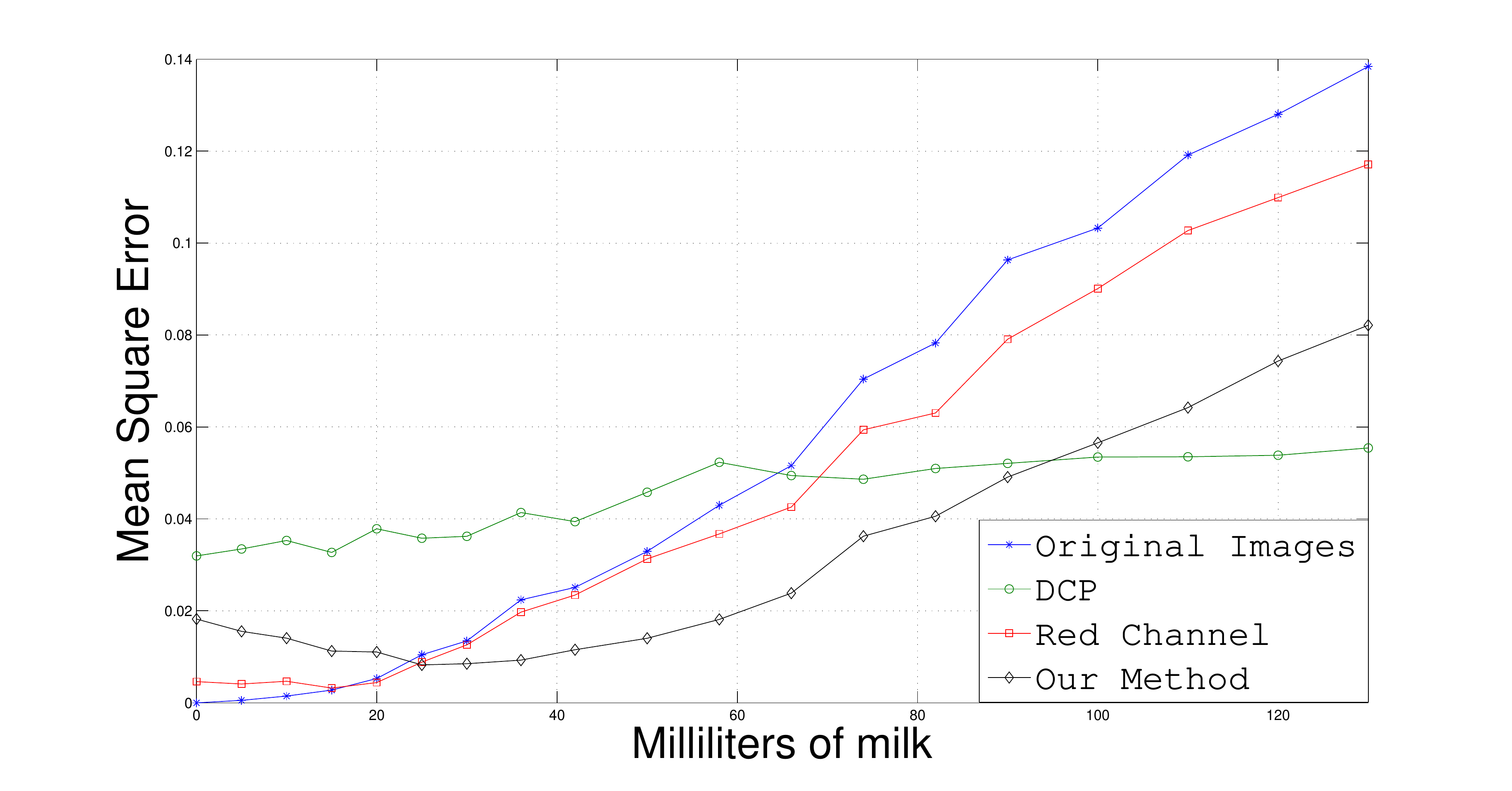}
\label{subfig:binf}
}
\subfloat[$A_\lambda^D$ estimated]{
\includegraphics[width=60mm]{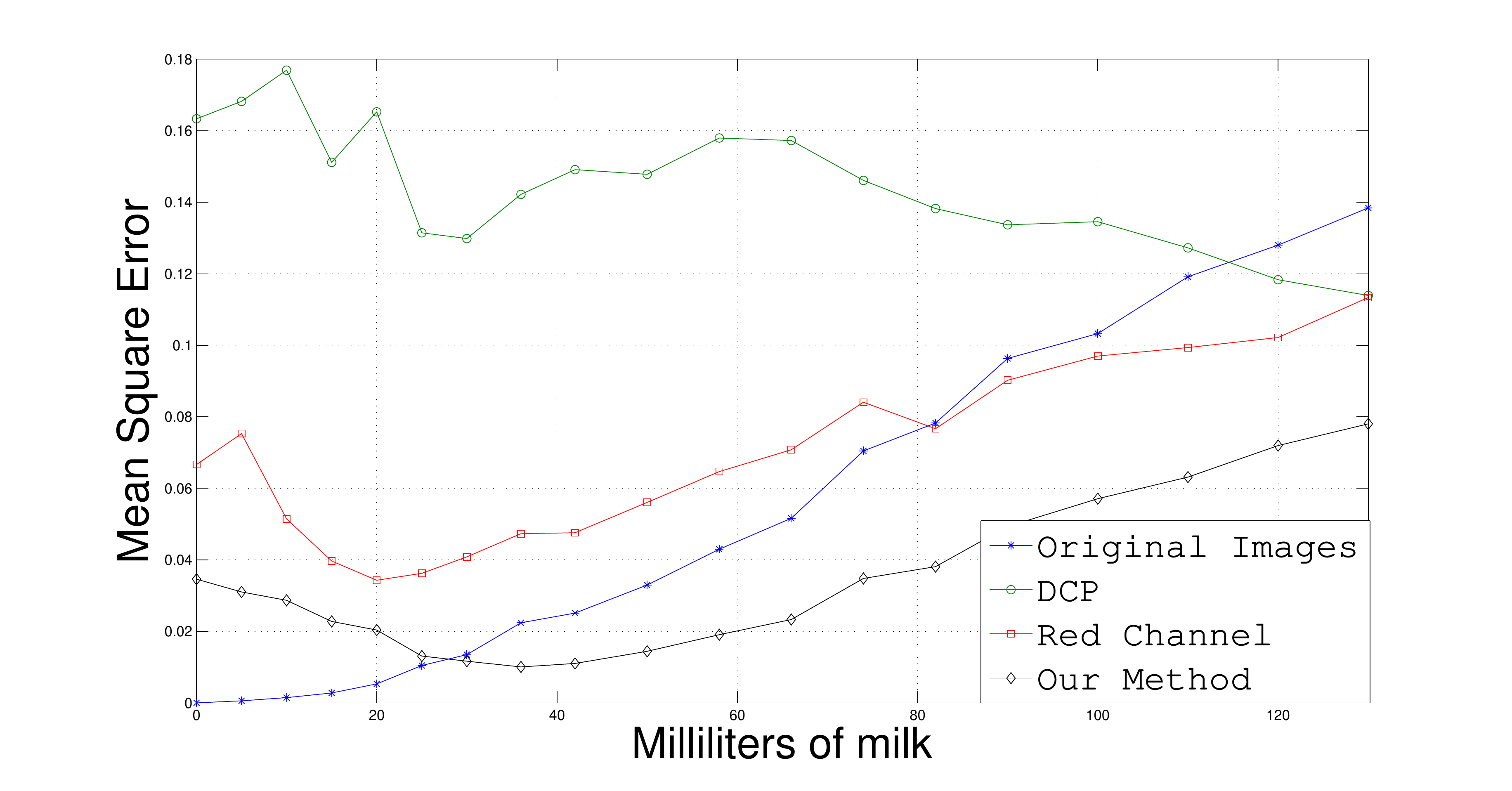}
\label{subfig:nobinf}
}
\caption{Quantitative evaluation of the proposed method using
the TURBID dataset \citep{duarte2016dataset}. The curves represents the MSE in function of quantity of milk. Each degraded image is compared with the clean image (Fig. \ref{subfig:clean}).}
\label{fig:plot}
\end{figure}

We show on Fig. \ref{fig:plot} the mean square error in function of the quantity of milk that is related with the degradation produced by the participating medium. The error is measured based on the difference between the restored image and the clean image (no milk). We compare our result with the Red Channel Prior (RCP) \citep{Galdran15}, the Dark Channel Prior (DCP) \citep{He09} and also the degraded images (original images), \textit{i.e.} without any kind of restoration. The blue line presents the error obtained using the degraded images. Based on the error of the degraded image, our method performs an effective restoration since the image becomes closer to the ground truth.

On Fig. \ref{subfig:binf}, the ambient light constant remains the same. In this case, the proposed method achieved the lowest average error, having a higher error than DCP on strongly degraded images. On Fig. \ref{subfig:nobinf}, each method estimates its own ambient light constant. The proposed method performed better when compared with the other methods since the ambient light constant estimated by DCP and the RCP are not accurate.

% Close Composite Transmission Estimation
%---------------------------------------------------------------------

%-------------------------------------------------------------------------
% Conclusion
%
\section{Conclusions}
\label{sec:conc}

In this paper we proposed a novel automatic single image restoration method to restore images captured in participating media. The method is inspired on the insufficiency of single priors restoration and their lack of generality for the use on multiple environments. We proposed two novel priors to be used in a larger variety of environments and a way to robustly integrate them. These priors and the image formation model produces an algorithm to be used on a large range of environment conditions. This generality is one of the main contribution of this paper.

We evaluated the proposed method in several kinds of participating medium obtaining competitive results related with the state-of-the-art. Finally, we also evaluated our method quantitatively by using the TURBID dataset. We observed that the proposed method is able to restore images in a larger range of degradation conditions. Finally, as a future work, we intend to assess several strategies to fuse transmission priors, and incorporate other source of information in our transmission estimation.

% Close Conclusion
%-------------------------------------------------------------------------

%---------------------------------------------------------------------
%Bibliography
%
\bibliographystyle{unsrtnat}
\bibliography{bibliography}
%---------------------------------------------------------------------
\end{document}